\documentclass[%
 reprint,
noeprint,
nofootinbib,
 amsmath,amssymb,
 aps,
prx,
floatfix,
]{revtex4-2}

\usepackage{graphicx}
\usepackage{dcolumn}
\usepackage{natbib}
\usepackage{mathptmx}      
\usepackage{url}            
\usepackage{booktabs}       
\usepackage{amsfonts}       
\usepackage{nicefrac}       
\usepackage{microtype}      

\usepackage{amsthm}
\usepackage{amsmath}
\usepackage{amssymb}
\usepackage{bm}
\usepackage{mathtools}
\usepackage{algpseudocode}
\usepackage{algorithm}
\usepackage{tikz}
\usetikzlibrary{calc}

\setcitestyle{numbers}
\setcitestyle{square}

\usepackage[caption=false, justification=raggedright]{subfig}
\usepackage[shortlabels,inline]{enumitem}

\usepackage{color}

\usepackage{multirow}
\usepackage{makecell}

\usepackage[hidelinks,colorlinks,bookmarksdepth=3]{hyperref}       
\usepackage{cleveref}
\usepackage[open]{bookmark}

\usepackage{pgfplots}
\pgfmathdeclarefunction{gauss}{2}{%
  \pgfmathparse{1/(#2*sqrt(2*pi))*exp(-((x-#1)^2)/(2*#2^2))}%
}
\pgfmathdeclarefunction{circle}{1}{%
  \pgfmathparse{2/(pi * #1^2) * sqrt(#1^2 - x^2)}%
}
\usepackage{transparent}

\usepackage{svg}

\usepackage{todonotes}

\usepackage{appendix}
\crefname{supp}{supp.}{supp.}
\Crefname{supp}{Supplement}{Supplements}

\crefname{assumption}{asm.}{assumptions}
\Crefname{assumption}{Asm.}{Assumptions}
\crefname{uassumption}{simplified asm.}{simplified assumptions}
\Crefname{uassumption}{Simplified asm.}{Simplified assumptions}
\crefname{lemma}{lemma}{lemmas}
\Crefname{lemma}{Lemma}{Lemmas}
\crefname{corollary}{corollary}{corollaries}
\Crefname{corollary}{Corollary}{Corollaries}
\crefname{algocfline}{algorithm}{algorithms}
\Crefname{algocfline}{Algorithm}{Algorithms}

\newcommand{\KL}{\text{KL}}

\newcommand{\ssystem}[1]{}
\newcommand{\bb}[1]{\mathbb{#1}}
\newcommand{\bv}[1]{\boldsymbol{#1}}

\newcommand{\overl}[1]{\overline{#1}}
\newcommand{\bvec}[1]{\boldsymbol{#1}}
\newcommand{\ca}[1]{\mathcal{#1}}

\newcommand{\dg}{\text{dg}}

\newcommand{\supps}{supp.\ }

\begin{document}


\title{Complexity from Adaptive-Symmetries Breaking:\\ Global Minima in the
Statistical Mechanics of Deep Neural Networks}

\author{Shawn W. M. \surname{Li}}
\email[Email: ]{shawnw.m.li@inventati.org}
\homepage[Homepage: ]{shawnwmli.ml}
\noaffiliation{}

\date{\today}

\newcommand{\details}[1]{#1}

\begin{abstract}
  \begin{description}
  \item[Background] The scientific understanding of complex systems and
    deep neural networks (DNNs) are among the unsolved important problems
    of science; and DNNs are evidently complex systems. Meanwhile,
    conservative symmetry arguably is the most important concept of
    physics, and P.W. Anderson, Nobel Laureate in physics, speculated that
    increasingly sophisticated broken symmetry in many-body systems
    correlates with increasing complexity and functional
    specialization. Furthermore, in complex systems such as DNA molecules,
    different nucleotide sequences consist of different weak bonds with
    similar free energy; and energy fluctuations would break the
    symmetries that conserve the free energy of the nucleotide sequences,
    which selected by the environment would lead to organisms with
    different phenotypes.
  \item[Purpose] When the molecule is
    very large, we might speculate that statistically the system poses in
    a state that would be of equal probability to transit to a large
    number of adjacent possible states; that is, an {\it adaptive
      symmetry} whose breaking is selected by the feedback signals from the
    environment. In physics, quantitative changes would accumulate into
    qualitative revolution where previous paradoxical behaviors are
    reconciled under a new paradigm of higher dimensionality (e.g.,
    wave-particle duality in quantum physics). This emergence of adaptive
    symmetry and complexity might be speculated as accumulation of
    sophistication and quantity of conservative symmetries that lead to
    a paradigm shift, which might clarify the behaviors of DNNs.
  \item[Results] In this work, theoretically and experimentally, we
characterize the optimization process of a DNN system as an {\it
extended symmetry-breaking process} where novel examples are
informational perturbations to the system that breaks adaptive
symmetries. One particular finding is that a hierarchically large DNN
would have a large reservoir of adaptive symmetries, and when the
information capacity of the reservoir exceeds the complexity of the
dataset, the system could absorb all perturbations of the examples and
self-organize into a functional structure of {\it zero training
errors} measured by a certain surrogate risk. In this diachronically
extended process, {\it complexity} emerges from quantitative
accumulation of adaptive-symmetries breaking.
  \item[Method] More
    specifically, this process is characterized by a
    statistical-mechanical model that could be appreciated as a
    generalization of statistics physics to the DNN organized complex
    system, and characterizes regularities in higher dimensionality. The
    model consists of three constitutes that could be appreciated as the
    counterparts of Boltzmann distribution, Ising model, and conservative
    symmetry, respectively:
    \begin{enumerate*}[(1)]
    \item a stochastic definition/interpretation of DNNs that is a {\it multilayer probabilistic
        graphical model},
    \item a {\it formalism of circuits} that perform biological computation,
    \item a {\it circuit symmetry} from which
      self-similarity between the microscopic and the macroscopic adaptability
      manifests.
    \end{enumerate*}
    The model is analyzed with a method referred as the {\it statistical
      assembly method} that analyzes the coarse-grained behaviors (over a symmetry
    group) of the heterogeneous hierarchical many-body interaction in
    DNNs.
  \item[Conclusion] Overall, this characterization
    suggests a physical-biological-complex scientific understanding to the optimization power of
    DNNs.
  \end{description}
\end{abstract}

\maketitle
\section{Introduction}
\label{sec:introduction}

In formulating the concept of organized complex systems
\cite{Scientific2016,Kornberger2003,Amderson1972,10.5555/1614219,erdi2007complexity,Nicolis:2012:FCS:2331101,thurner2018introduction},
P. Anderson, described in the seminal essay {\it More is Different}
that, ``it is only slightly overstating the case to say that physics
is the study of symmetry'' \cite[p. 393]{Amderson1972}, and in the
same essay, he prefigured \cite[p. 396]{Amderson1972}: ``... at some
point we harre to stop talking about decreasing symmetry and start
calling it increasing complication ...'' in ``... functional structure
in a teleological sense ...''\details{---teleology refers to
goal-directed behaviors controlled by feedback \cite{Rosenblueth1943},
and in modern terminology, complication is referred as complexity.}
\details{Organized complex systems are adaptive
systems emerge from a large number of locally nonlinearly interacting
units and cannot be reduced to a linear superposition of the
constituent units.} Despite the progress, the study of organized complex
 systems (e.g., biotic, neural, economical, and social
systems) is still a long way from developing into a science as solid
as physics \cite{Keller2009}. And the field has been considered to be
in a state of ``waiting for Carnot''; that is, the field is waiting
for the right concepts and mathematics to be formulated to describe
the many forms of complexity in nature \cite[p. 302]{10.5555/1614219}.

This tension between seemingly incomprehensible complexity and inquiry
of science also underlies the field of Deep Neural Networks (DNNs)
\cite{LeCun2015}. DNNs have solved marvelous engineering problems
\cite{Krizhevsky2012,He,Radford}, and posit to contribute to societal
challenges \cite{Rolnick2019} and difficult scientific problems
\cite{Senior2020}, but could also induce disruptive societal changes
\cite{Automation2019,deepfake}. However, the theoretical understanding
of DNNs is rather limited \cite{Plebe2019} for the same reason of that
of complex systems: a DNN is an organized complex system
hierarchically composed by an indefinite number of elementary
nonlinear modules. The limit in understanding manifests in the
optimization properties \cite{Glorot2010,Dauphin2014,Baity-Jesi2018}
and generalization ability \cite{Krizhevsky2012,Oquab2014,Zhang2016b},
and results in critical weakness such as interpretability
\cite{Lipton2018b,BarredoArrieta2020}, uncertainty quantification
\cite{Ovadia2019,Abdar2021}, and adversarial robustness
\cite{Szegedy2013,Goodfellow2018}, harbingering the next wave of
Artificial Intelligence \cite{Jenkins2020}.

This situation is not unsimilar to the state of physics at the turn of
the 20th century \cite{Carleo2019,Zdeborova2020}. In the edifice of Classic Physics,
a few seemingly incomprehensible phenomena questioned the fundamental
assumptions of contemporary physics (i.e., black-body
radiation and photoelectric effect), and catalyzed the development of
Quantum Physics. When a field of science sufficiently mature in the
sense of explaining all existing phenomena, the reconciliation of
paradoxical properties of a system in the existing science with a new
``science'' is summarized by Thomas S. Kuhn \cite{Kuhn:1970} as {\it
paradigm shift} (e.g. wave-particle duality in quantum
physics, and the mass-energy equivalence in special relativity). And a
commonality of the paradigm shifts is that quantitative change
accumulates into qualitative revolution, and the previous
paradoxical behaviors are resolved by a model of higher
dimensionality: the phenomena reduced to quantum scale
need quantum physics, and are characterized not by point mass as a
particle or a basis function (which is an infinitely long vector) of a
particular frequency as a wave, but the probability of being in
certain states that are eigenvectors of a matrix\details{; and the phenomena at
the speed of light need the special theory of relativity, and are
characterized by a spatial-temporal four dimensional model that
incorporates the speed of light, instead of the spatial three
dimensional model}.

In this work, we develop a characterization of the optimization
process of DNNs' that could be appreciated under this concept of
paradigm shift; that is, a DNN being a statistical-mechanical system,
as the sophistication and quantity of symmetries increases
quantitatively, the statistical mechanics of this DNN organized
complex system needs to and could be characterized by concepts of
higher dimensionality than the ones in the statistical physics of
disorganized complex systems. This characterization suggests a
scientific---in the sense of epistemology and methodology of physics
\cite{Scientific2016,Laughlin2000,entropy,bailly2011mathematics,Longo2011,Longo2013},
and of the philosophy of falsifiable science
\cite{Popper:34}---understanding of the optimization process and thus
the optimization power of DNNs; a particular interesting result is that,
informally stated, global minima of DNNs could be reached when the
potential complexity of the DNN (which could be enlarged by making the
DNN hierarchically large) is larger than the complexity of the
dataset.

The formal characterization of this process would require us to visit
the epistemological foundation of physics
\cite{Laughlin2000,Laughlin2000a,Scientific2016,entropy,bailly2011mathematics,Longo2011,Longo2013},
to extend biologists and complexity scientists' revision of this
foundation
\cite{10.5555/1096919,Scientific2016,Kauffman1991,Davidson2006,Davidson2010,Li2010b,Krakauer2011a,bailly2011mathematics,Longo2011,Longo2013,Montevil2016,Mossio2016,Flack2017a,Wolf2018,Ramstead2018,Jost2021a},
and to leverage on recent progress in probability and statistics
\cite{Helton2007,wainwright2008graphical,Erdos2017,Alt2018a}. Furthermore,
we need to come up a {\it statistical-mechanical model of the DNN
organized complex system} consisting of three constituents that could
be appreciated as the counterparts of Boltzmann distribution, Ising
model, conservative symmetry, respectively of statistical-physical
models. The resulted model is analyzed with a method referred as the
{\it statistical assembly method} that analyzes the coarse-grained
behaviors (over a symmetry group) of the heterogeneous hierarchical
many-body interaction in DNNs.

Therefore, to increase readability, this work is presented in a
fractal style: we have written a short letter \cite{Shawn2021letter} to give a high level
overview of the results, and it is advised to read the letter first;
in this article, we shall elaborate the overview there.
To begin with, we shortly summarize the main messages in the letter
\cite{Shawn2021letter} in the next paragraphs.

By discussing the epistemology of physics, where the
renormalizibility of physical systems
\cite{McKay1982,ROSATI2001,Caglar2017} is an emergent organizing
principle founding on {\it conservative symmetries}
\cite{Amderson1972,Longo2019} in which the microscopic details are
unknowable or irrelevant \cite{Laughlin2000}, we motivate that we
might need to investigate mesoscopic organizing principles of organized
complex systems \cite{Laughlin2000a} that are stable
\cite{Laughlin2014b} from an informational perspective
\cite{Krakauer2011a,Flack2017a,Seoane2018,Krakauer2020,Ramstead2018}. The
proliferation of adjacent states
\cite[p. 263]{thurner2018introduction} with similar free energy (i.e.,
conservative symmetric states) in DNA macromolecules \cite{Wolf2018},
and the transitions among those states by the breaking conservative
symmetries (induced by spontaneous symmetry breaking in quantum field
theories or deterministic chaos \cite{Bravi2015,Longo2019}) make us speculate
that the lack of conservative symmetries in DNNs
\cite{Mehta2014,Lin2017} is a feature looked in a backward
perspective.

The symmetry breaking in biology \cite{Li2010b,Longo2011,Flack2012}
breaks the symmetry of adaptation: the system has the capacity to process
the novel information---that is, to adapt---by posing in states where
symmetric possible directions to adapt could be adopted, which is in
turn induced by the complex cooperative interaction among the
heterogeneous units in a biotic system; and the symmetric states would
break in response to random fluctuations and external feedback signals
\cite{Li2010b}.  Thus, increased sophistication and quantity of
conservative symmetries might lead to an antithetical concept of {\it
adaptive symmetry}: unlike the symmetries in physics, which formalizes
a conservative law that conserves the free energy of different states
related by certain transformations and thus characterizes the
invariant of free energy, the adaptive symmetry is the conservation of
\textit{change of free energy}; or in other words, the invariant of change
that emerges as a result of the increased sophistication and quantity
of conservative symmetries. Furthermore, in a complex adaptive
system, a microscopic change at one scale has implications that ramify
across scales throughout the system \cite{Levin1997,Krakauer2011}, and
thus regularities are not to be found in coarse-grained static
behaviors where higher-order coupling is considered irrelevant
fluctuations, but in the dynamic behaviors of adaptation
\cite{Longo2011,Montevil2016,Mossio2016,Jost2021a}.

Motivated by the
speculation, we investigate and find self-similar phenomena in DNNs
where the output of a DNN (macroscopic behaviors) is the
coarse-graining of basis units composed by neurons (that are referred as
\textit{basis circuits}, and are microscopic behaviors), and both the
DNN and the basis circuits are dynamical feedback-control loops
\citep{Rosenblueth1943,10.5555/1096919,Heylighen2001} (between the
system and the environment) that are of adaptive symmetry. This
self-similarity is concentration-of-measure phenomena that are of
higher intrinsic dimensionality than those of disorganized complex
physical systems. And complex functional structure emerges in a
diachronic process of adaptive-symmetries breaking in response to
feedback signals. Furthermore, during the process, intact and broken
adaptive symmetries coexist---the former maintains the adaptability of
the system, and the latter manifests as functional structure. Thus,
for a hierarchically large enough DNN (relative to the dataset), a
phenomenon exists such that sufficient adaptive-symmetries enable
convergence to global minima in DNN optimization, and this process is
an extended symmetries-breaking process that is both a phase and a
phase transition---ringing a bell of paradigm shift.

In this article, we elaborate the results discussed only briefly in
the letter \cite{Shawn2021letter}. In the rest of the
introduction, we shall give the outline.
And in the main body we shall mostly discuss the theoretical results
conceptually at an intuitive level of formalism along with
experimental results. We also have a formal presentation from an
axiomatic approach in the sense of Hilbert's sixth problem
\cite{Gorban2018}; but the rigor and heavy math in this style are
likely only interesting to a small subset of the audience, and thus
the formal version of the main results is given in the supplementary
material \cite{Shawn2021supparxiv}.

\subsection{Outline}
\label{sec:outline}

\subsubsection{Statistical-mechanical model of DNN complex system}
\label{sec:outline-main-body}

The constitutes of the statistical-mechanical model of DNNs is
introduced from \cref{sec:umwelt} to \ref{sec:adapt-symm-feedb}. The
constituents are introduced succinctly as follows.

\begin{enumerate}[leftmargin=0cm]
\item First, in \cref{sec:umwelt}, we introduce a system that does
statistical inference on hierarchical events, which is contextualized
as a measure-theoretical formalism of the concept {\bf Umwelt} in
theoretical biology. More specifically, a statistical characterization
(i.e., {\bf Boltzmann distribution}) of the behaviors of disorganized
complex systems is possible because despite the unpredictable
behaviors of individual units, these irregularities are subdued into
predictable aggregate behaviors of a vast number of units that are
subject to average-energy constraints. Complex biotic systems also
have such mixture of irregularities and predictability: evolution of
biotic systems are both the result, of random events at all levels of
organization of life and, of the constraint of possible evolutionary
paths---the paths are selected by interaction with the ecosystem and the
maintenance of a possibly renewed internal coherent structure of the
organism that are constructed through its history
\cite{Longo2013}. The proposed Umwelt is a statistical-inference
system that models this phenomenon under the context of DNNs. Compared
with the disorganized complexity in statistical physical, where a
system maximizes entropy subject to the constraint that system's
energy (i.e., the average of energy of units) is of a certain value,
an Umwelt organized-complex system maximizes entropy subject to
hierarchical event coupling such that it estimates the probability
measures of groups of events in the environment that forms a
hierarchy, and the probability measure of certain coarse-grained
events with fitness consequences are estimated and could be used to
predict future events.

\item Second, in \cref{sec:bayes-defin-dnns}, we introduce a {\bf
stochastic definition of DNNs}, that is an implementation of the
Umwelt---through a hierarchical parameterization scheme that estimates
probability measures (for the statistical inference problem of Umwelt)
by dynamical programming, and an approximation inference method---and is
a supervised DNN with the ReLU activation function. The definition
defines a DNN as a {\bf multilayer probabilistic graphical model} that
is a hybrid of Markov random field and Bayesian network. The
definition could be appreciated as a sophisticated deep higher-order
Boltzmann machine with priors on weights, though there are critical
differences, which is discussed in \supps A A with
related works for interested readers.
\item Third, in \cref{sec:dnn-self-organ}, motivated by biological
circuits, we introduce a {\bf formalism of circuits} upon the
stochastic definition of DNNs that formalizes the risk minimization of
a DNN as a {\bf self-organizing} informational-physical process where
the system self-organizes to minimize a potential function referred as
{\bf variational free energy} that approximates the entropy of the
environment by executing a {\bf feedback-control loop}. The loop is
composed by the hierarchical {\it basis circuits} (implemented by
coupling among neurons) and a {\it coarse-grained} random variable (of
fitness consequences) computed by the circuits. And each basis circuit
is a microscopic feedback-control loop, and the coarse-grained effect
of all the microscopic loops computes the coarse-grained
variable. Under this circuit formalism, a basis circuit is like a spin
in the {\bf Ising model}; for example, the order parameter of DNNs
derived from the circuit formalism is symbolically (but only
symbolically) equivalent to the spin glass order parameter---order
parameter of DNNs is introduced in \cref{sec:order-from-fluct}.
\item Fourth, in \cref{sec:adapt-symm-feedb}, we introduce an {\bf
adaptive symmetry}, referred as {\it circuit symmetry}, of the basis
circuits that characterize the phenomenon that each basis circuit is
of a symmetric probability distribution, and thus of equal probability
to contribute to the coarse-grained variable positively or
negatively. The symmetry could break to a positive or negative value
in response to feedback signals, and thus reduces risk. The symmetry
is a {\it heterogeneous symmetry} (which shall be further explained in
\cref{sec:adapt-symm-feedb}), implying that intact and broken circuit
symmetries could coexist in a DNNs: the broken circuit symmetries
encodes information about the environment (i.e., datasets), and the
intact circuit symmetries maintain the adaptability to further reduce
informational perturbations (i.e., training examples with nonzero
risk).  Adaptive symmetries are not symmetries typically in physics
that conserve systems' free energy (which is referred as {\bf
conservative symmetries} in this work). This epistemological
difference is introduced in \cref{sec:from-cons-symm}. To analyze the
adaptive symmetries, a synthesis of {\bf statistical field theory} and
{\bf statistical learning theory} is developed, and is referred as
{\bf statistical assembly methods}. The method analyzes the
coarse-grained behaviors of basis circuits summed over symmetries
group that turn out to be self-similar to the behaviors of basis
circuits---which is similar to what the renormalization group technique
reveals for the coarse-grained effect of symmetries in physics.
\end{enumerate}

\subsubsection{Extended symmetry breaking of DNNs}
\label{sec:outline-extended-symmetry-breaking}

The symmetry breaking process of DNNs is introduced from in
\cref{sec:order-from-fluct} and \ref{sec:plast-phase-benign}. We also
lightly discuss the paradox under the context of complexity science
that the adaptive-symmetries breaking processing is both a phase and a
phase transition in \cref{sec:disc-compl-from}. The key messages of
the sections are summarized as follows.

\begin{enumerate}[leftmargin=0cm]
\item In \cref{sec:order-from-fluct}, we introduce the {\bf order} of
DNNs, which generalizes the order in statistical physics (e.g., the
magnetization is the order of a magnet) and the order in
self-organization (i.e., order from noise/fluctuations/chaos), and
characterizes the coarse-grained effect of circuit symmetries in a DNN
to continue absorbing informational perturbations; that is, the
capability of a DNN to decrease risk.

\item In \cref{sec:plast-phase-benign}, we introduce a phase of DNNs,
which is referred as the {\bf plasticity phase}. In this phase, while
a DNN continually breaks circuit symmetries to reduce informational
perturbations, the large reservoir of circuit symmetries result in
that circuit symmetries stably exist (along with broken circuit
symmetries), and the system manifests a self-similarity between the
basis circuit (i.e., microscopic feedback-control loop) and the
macroscopic/coarse-grained adaptive symmetry at the level of neuron
{\it assemblies} (i.e., macroscopic/coarse-grained feedback-control
loop). Both the gradient and Hessian are computed by neuron
assemblies, and thus are of the adaptive symmetry; more specifically,
the both the gradient, and eigenspectrum of Hessian are of a symmetric
distribution---a subtlety exists in the problem setting of this work,
and strictly speaking it is not that the eigenspectrum of Hessian is
symmetric but a matrix in a decomposed form of Hessian. The phase is
both a phase of a DNN, where the DNN could continually decreases risk,
and an {\bf extended symmetry-breaking} process, where the symmetries
are continually being broken during the self-organizing process.  As a
result, all stationary points are saddle points, and benign pathways
on the risk landscape of DNNs to {\bf zero risk} could be found by
following gradients. Overall, the results suggest an explanation of
the optimization power of DNNs.
\end{enumerate}

The training of a DNN is both an extended symmetry-breaking process,
and a plasticity phase with stable symmetries. This superficially
looks like a paradox: in physics, symmetry breaking is usually a
singular phase transition. However, the unification of previously
paradoxical properties of a system/object has repeatedly happened in
the history of physics: at the quantum scale, wave and particle become
a duality in quantum physics, and at the speed of light, the mass and
energy become equivalent in special relativity. This is referred as
the paradigm shift by Thomas S. Kuhn \cite{Kuhn:1970}. Thus,
\cref{sec:disc-compl-from} aims to address possible confusions by
outlining a very crude look at the whole of the symmetry-breaking
process of DNNs, and relating this work to complexity science
\cite{Kornberger2003,Amderson1972,10.5555/1614219,erdi2007complexity,Nicolis:2012:FCS:2331101,thurner2018introduction}.

\paragraph{Discussion.}
More particularly, the increased sophistication and quantity of
symmetries of biotic systems makes symmetry-breaking gradually
transits from a singular process to a continual process, and the
plasticity phase could be understood as a diachronic process of
evolving: when upper limit/bound of the complexity of the system is
larger than the complexity of the environment where the system
embodies, the system could perfectly approximate the
organization/entropy of the environment (measured in surrogate
risk). And the self-organizing process of DNNs is summarized as {\bf
complexity from adaptive-symmetries breaking}, which characterizes a
process where a system computes to encoded increasing complex
information by breaking adaptive symmetries.

\subsubsection{Problem setting}
\label{sec:problem-setting-intro}

\paragraph{Theoretical setting.}
To conclude the introduction, we summarize the setting of the
theoretical characterization in \cref{sec:problem-setting}.  Einstein
said in a lecture in 1933, ``it can scarcely be denied that the
supreme goal of all theory is to make the irreducible basic elements
as simple and as few as possible without having to surrender the
adequate representation of a single datum of
experience''\cite{Robinson2018}. We identify such a minimal
irreducible DNN system that roughly is a hierarchically large
deep/multilayer neural networks with ReLU activation function and a
feedforward architecture, doing binary classification on real-world
data whose complexity is less than the potential complexity of the
network. As a clarifying note, we have mentioned in
\cref{sec:outline-extended-symmetry-breaking} in the passing that a
subtlety exists in the symmetry of Hessian's eigenspectrum. The
subtlety comes from the class of loss functions that this paper works
with, which is introduced in \cref{sec:problem-setting}.

\paragraph{Experimental setting.}
We also validate theoretical characterizations with experiments as the
narrative develops. In the experiments, we train a VGGNet
\cite{Simonyan2014} (cross entropy loss replaced by the hinge loss) on
the CIFAR10 datasets modified for binary classification. The DNN has
$12$ layer, $\sim 10^7$ parameters/weights. More details of the
experiment setting is given in \supps H B. The experiments
should be understood as simulations that validate the theoretical
characterizations. And to appreciate the simulation, existing works
that does finite-size correction through statistical field theory to
the works that study DNNs in the infinite-width setting tend to
validate theoretical characterizations by running on toy models
processing toy, or unrealistic datasets; for example,
\citet{Cohen2019} test the theory with four-layer large-width MLP on
data uniformly sampled from a hypersphere, and justify their
simplification by stating that expecting analytical characterization
of networks of VGG architecture on ImageNet would be to expect ``an
analytical computation based on thermodynamics to capture the
efficiency of a modern car engine or one based on Naiver-Stoke’s
equations to output a better shape of a wing
\cite[p. 11]{Cohen2019}''; further contexts on existing works
motivated by statistical field theory could be found in
\supps A D. Though we only run simulation with a
VGG network on a classic small dataset CIFAR10, this should be
considered as a supportive experimental validation of the theoretical
characterizations, and a start towards industrial settings,
considering the difficulty of the problem.

\subsubsection{Related works}
\label{sec:outl-relat-works}

The narrative of this work resolves around the characterization that
the optimization of DNNs is an extended symmetry-breaking, and ends at
an explanation of the optimization power of DNNs. Meanwhile, the
narrative is composed by a stochastic definition/model of DNNs, a
circuit formalism that analyzes the model, an epistemologically
different symmetry (i.e., adaptive symmetry) with the conservative
symmetry of physics, and the study of order and a phase of DNNs
through the so call statistical assembly methods. Each of these
components have extensive related works on its own, except for the
circuit formalism, which should be appreciated as a technique that
analyzes and accompanies the statistical definition. Therefore, to put
the resultts under more specialized context, in addition to the
background discussed extensively in the main body, we also collect
discussion on these related works separately in
\supps A for interested readers.

More specifically, first, the stochastic definition of DNNs in this
work is a stochastic DNN, and also a Bayesian DNN, and thus existing
works that study stochastic neural networks and Bayesian neural
networks are discussed in \supps A A. Second,
existing works that interpret the operation of DNNs as circuits are
discussed in \supps A B, which is very short because they
are only remotely related. Third, existing works that study symmetries
of DNNs are discussed in \supps A C. Fourth, related
works that study the optimization and phases of DNNs are discussed in
\supps A D, by putting this works under the
context of related works that study the risk landscape under the {\bf
overparameterized} regimen, and the phases of DNNs.

\subsection{Notations}
\label{sec:notations}

Normal letters denote scalar (e.g., $f$); bold, lowercase letters
denote vectors (e.g., $\bvec{x}$); bold, uppercase letters denote
matrices, or random matrices (e.g., $\bvec{W}$); normal, uppercase
letters denote random elements/variables (e.g., $H$). $\dg(\bv{h})$
denotes a diagonal matrix whose the diagonal is the vector
$\bv{h}$. $:=$ denotes the ``define'' symbol: $x := y$ defines a new
symbol $x$ by equating it with $y$.  A sequence of positive integer is
also conveniently denoted as $[N] := \{1, \ldots, N\}$. $\bb{B}$ denotes
$\{0, 1\}$. The upper arrow on operations denotes the direction: for
example, $\overrightarrow{\Pi}_{i=1}^{n}\bv{W}_i,
\overleftarrow{\Pi}_{i=1}^{n}\bv{W}_i, i < n, i, n \in \bb{N}$ denote
$\bv{W}_1\ldots\bv{W}_n, \bv{W}_n\ldots\bv{W}_1$, respectively. $\bv{i}_{p:q}$
denotes the subvector of $\bv{i}$ that is sliced from the $p^{th}$
component to $q^{th}$ (exclusive; that is, $i_q$ is excluded). This is
the conventional in most programming languages to slice arrays. If the
ending index is omitted, e.g., $\bv{i}_{p:}$, it denotes the subvector
sliced from the $p^{th}$ component until the end (inclusive);
similarly, if the starting index is omitted, e.g., $\bv{i}_{:q}$, it
denotes the subvector sliced from the beginning (inclusive) until the
$q^{th}$ component (exclusive).  Because we shall deal with random
variables in a multilayer network, we need the indices to denote the
layers. When the lower index is not occupied, we use the lower index
to denote a layer; for example, random vector at layer $l$ is denoted
$H_l$. Otherwise, we put the layer index onto the upper index; for
example $H^{l}_i$ denotes $i^{th}$ component of $H_l$. If matrices are
indexed, we move the index up when indexing its entries, i.e.,
$w^1_{ij}$ denotes the $ij$th entry of $\bv{W}_1$.  The other
notations should be self-evident.

\section{Main results}
\label{sec:main-body}

\subsection{Umwelt: system that does statistical inference on hierarchical events}
\label{sec:umwelt}

\subsubsection{Boltzmann distribution, disorganized complexity, complex biotic systems and DNNs}
\label{sec:boltzm-distr-disorg}

\begin{figure*}[t]
  \centering
  \includegraphics[width=\linewidth]{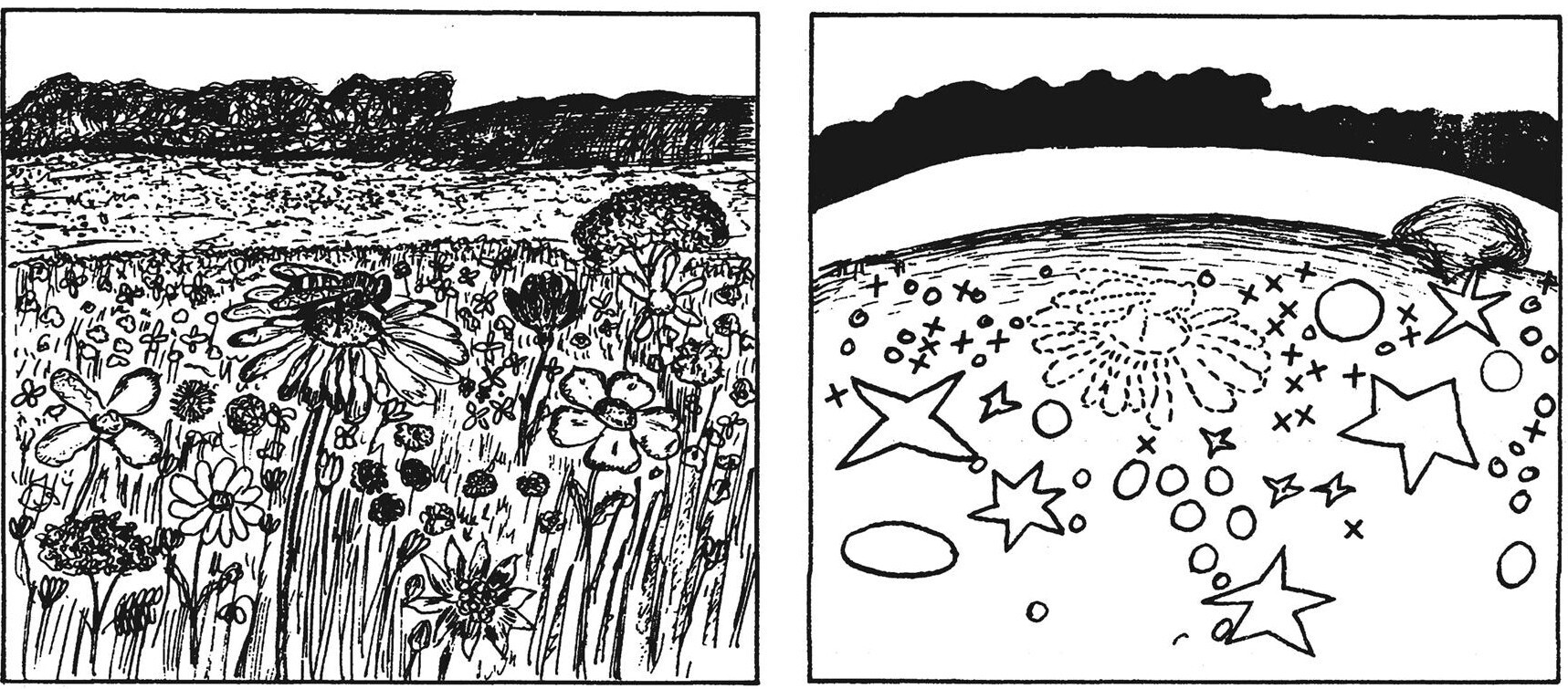}
  \caption{Environment and Umwelt of a honey bee, illstration from
\citet[p. 351]{Uexkull199}. On the left is the environment observed by
an external observer. On the right is the Umwelt of honey bees, which
consists of the events (observations of certain objects) that have
effects on the honey bees (e.g., the flowers with nectar), and bees
could have an effect on (e.g., gathering the nectar). In this work, we
formalize the emergence of an Umwelt as a statistical-inference
process that estimates the probability measures of groups of events in
the environment that forms a hierarchy, such that the probability
measure of certain coarse-grained events with fitness consequences
(e.g., detection of flowers with nectar) are estimated and could be
used to predict future events. The probability measures are estimated
through hierarchical entropy maximization subject to constraints that
characterize hierarchical event coupling. With a hierarchical
parameterization scheme that estimates probability measures through
dynamical programming, and a particular choice of approximated
inference method, the probability-estimation/learning algorithm of
Umwelt is an expectation-maximization algorithm that corresponds to
the forward-backward supervised training of a DNN with ReLU activation
function. This formalism leads to a stochastic (whitebox) definition
of DNNs that is a multilayer probabilistic graphical model.}
  \label{fig:umwelt}
\end{figure*}

In the 19th century, Ludwig Boltzmann provided a statistical mechanic
model of gases at thermal equilibrium that provided a causal
characterization of macroscopic thermodynamic behaviors of gases from
microscopic gas molecules. The model characterizes a phenomenon that
could be informally stated as: the gas in a container consists of
billions of atoms that would predominantly stay close to certain
states dictated by the system's energy, and their chance of being in
other states decreases exponentially; the probability distribution in
the model is known as the {\bf Boltzmann distribution}. Despite the
exactness and clarity now conveyed by the model, the model was
proposed among deep philosophical confusions, which could be
summarized by the great debate between Ernst Mach and L. Boltzmann:
``Boltzmann needed atoms for his thermodynamics, and he wrote a
passionate plea titled {\it On the Indispensability of Atoms in
Science}. However, since atoms cannot be directly perceived, Ernst
Mach treated them as mere models, constructions of the mind, not all
that different from his old bugaboo—Kant’s Thing-in-Itself. Whenever
atoms were mentioned, Mach would slyly ask, with a clear Viennese lilt
in his voice: `Did you ever see one?'''
\cite[Chp~2]{sigmund2017exact}. Although now the current technology
enables us to directly observe atoms, the philosophical problem of
observables simply has been pushed down to quantum physics, which
relies on a concept of {\it effect theory} \cite{Laughlin2000}.  Such
ambiguity in the fundamental concepts have been major obstacles in
developing scientific theories in physics \cite{ETJaynes1967}---we shall
discuss the epistemology of physics as this paper develops.

From the perspective of history of science, the study of the
mechanical causality between neurons and cognition (or more generally,
mind) is also under this tension between microscopic neurons and
mathematical models of cognition.  The interaction among a population
of neurons of an organism are informational, depends on inputs,
context and history of the organism, and thus models of neuron
population typically characterize episodes of the behavior of the
neurons where prepackaged information is fed to the neurons, and the
neuron behaviors are interpreted and analyzed through observables
identified by a mathematical model. And thus different models with
different input information would give different mechanisms of how
microscopic neurons lead to macroscopic cognition---or do not consider
cognition as a macroscopic phenomenon of neurons at all. For example,
it is still not clear whether neurons transit information by
modulation of their mean firing rate, or spiking-time dynamics
orchestrated by a population of neurons \cite{Eguchi2018}; and the
relationship between neurons/brain and mind is still philosophical
speculations \cite{Lopez-Rubio2018}. And the role of individual
neurons in DNNs is not well understood, and under various hypotheses,
existing works have proposed various methods to visualize the
information encoded by neurons
\cite{Mahendran2014,Raghu2017,Bau2020,Hoffmann2021}. Such ambiguity in
the basic concept of a neuron has prevented scientific analysis of
DNNs, and resulted that DNNs are considered blackboxs
\cite{Castelvecchi2016}.

A statistical mechanic model of gases is possible because disorganized
complex systems (e.g., gases) has the characteristics that although
the behaviors of unit, or a small group of units are irregular and
unpredictable, the irregularities are subdued into the aggregated
behaviors of a vast number of units that are subject to average-energy
constraints and thus predictable and amenable to analysis. However,
complex biotic systems also have such mixture of irregularities and
predictability. Evolution of biotic systems are both the result, of
random events at all levels of organization of life, and of the
constraint of possible evolutionary paths---the paths are selected by
interaction with the environment and the maintenance of a possibly
renewed internal coherent structure of the organism that has been
constructed through its history \cite{Longo2013}. For example, the
beak of Darwin's finches are adapted to the different environments of
islands, and the adaptation is also constrained by the previous shape
of the ancestry's beak. Therefore, in the study of complex
adaptive/biotic systems, it is instructive to identify a hierarchy of
increasingly constrained models based on the adaptive properties
\cite{Krakauer2010}.

Under this context, to analyze DNNs, we design a
probability-estimation, or colloquially, learning system that does
statistical inference on hierarchical events; and an implementation of
the system---which includes the choice of parameterization and
approximated inference methods---would be a supervised DNN with the ReLU
activation function. Compared with the disorganized complexity in
statistical physics, where a system maximizes entropy subject to the
constraint that system's energy (i.e., the average of energy of units)
is of a certain value, the learning system maximizes entropy subject
to hierarchical event coupling that encodes regularities in the
environment. We introduce the system in the following---a more rigorous
formalism is given in \supps B B.

\subsubsection{The Umwelt statistical-inference system, and its biological motivation}
\label{sec:biol-motiv-umwelt}

Evolution of biotic systems by natural selection
\cite{darwin-originofspecies-1936} could be interpreted
information-theoretically as an {\it inferential process}
\cite{Krakauer2011a,Flack2017a,Seoane2018,Krakauer2020,Ramstead2018}
where a system minimizes the uncertainty of its hypotheses on the
environment. More specifically, the {\it genotype} (all the DNA
molecules) could be the states of the system. The genotypes of biotic
systems encode hypotheses \cite{Koonin2016} about the present and
future environments that guide the systems to perpetuate themselves in
these environments. Evolution is an inferential process, where a
system evolves to minimize the uncertainty of the hypotheses, such
that when a system interacts with the environment, the predicted
consequence of its action would be close to the realized consequence
with high probability instead of leading to uncertain outcomes. The
effective hypotheses are selected by the environments and are passed
onto future generations.

Furthermore, a biotic system adapts to and occupies an ecological
niche \cite{Krakauer2011a} of an environment, and thus it has been
hypothesized that it only builds a model of the niche, instead of the
whole ecosystem. The hypothesis has been conceptualized as {\it
Umwelt} \cite{Uexkull199}, which denotes all aspects of an environment
that have an effect on the system and can be affected by the system
\cite{Ay2015a}. An insightful illustration of the concept from
\citet{Uexkull199} is given in \cref{fig:umwelt}.

The learning setting in statistical learning theory
\cite{mussardo2020statistical} could be appreciated as a simplistic
formalism of the interaction between adaptive systems and an
environment: a system is formalized as a hypothesis in a hypothesis
space, and the dataset is the environment. Formally, the {\bf observed
environment} is a tuple $(Z, S_m)$ that consists of a random variable
$Z := (X, Y)$ with an unknown law $\mu$, and a sample $S_m = \{ z_i =
(\bvec{x}^{(i)}, y^{(i)}) \}_{i \in [m]}$ of size $m$ (i.e., observed
events). In the following, we shall design the hypothesis space.

In the statistical inference problem of a Umwelt, an exponential
number of combinations of events are possible in an environment, and
thus inference of future events is possible only if regularities exist
in the combinations. To appreciate the statement, we could look at
regularities in physics. A fundamental regularity in physics is
symmetries: a gram of magnet contains enough molecules/atoms such that
if each combination of molecules behaves differently, the number of
possible behaviors of the combination would exceed the number of atoms
of the universe, and thus makes it impossible to make statistical
prediction because there is no stable population behaviors and an
identical copy of the crystal is needed to make predictions; however,
the behaviors of the molecules at different spatial locations are
highly homogeneous---i.e., translational symmetry---such that the
average/statistical behavior of the whole could be encoded by a single
macroscopic ``spin'', which is referred as a {\it field} in
statistical field theory \cite{mussardo2020statistical}.

To investigate regularities in an environment, we hypothesize that the
future events are inferred from coarse-grained events that are formed
by a hierarchy of groups of events, and are relevant for the biotic
system's fitness, and thus the emergence of a Umwelt is formalized as
a statistical-inference process that estimates the probability
measures of groups of events in the environment that forms a
hierarchy, such that the probability measure of certain coarse-grained
events with fitness consequences are estimated and could be used to
predict future events.

More specifically, the formalism of {\bf Umwelt} is a
recursive/self-similar system that could be further appreciated
through the concepts of {\it signals} and {\it boundaries}
\cite{holland2012signals}, as introduced in the following.
\begin{enumerate}[leftmargin=0cm]
\item A biotic system has its own internal dynamics enclosed by a
boundary that receives signals from the environment and acts upon the
environment. For example, a neuron cell is enclosed by membranes
(which consists of specialized molecules that exchange signals with
the environment), intakes neurotransmitter from other neurons, but
also outputs neurotransmitters to other neurons. And thus, the system
occupies a niche in the environment and only interacts with the subset
of signals among all signals. Such an internal dynamics could be
modeled probabilistically through a Markov probability kernel $Q(a,
s)$ (where $a$ and $s$ are numerical representations of actions and
signals, respectively) that formalizes the phenomenon that a specific
signal $s$ would elicit an action of the system with probability $Q(a,
s)$. Let $S$ denote the random variable whose realizations are actions
$s$. By conditioning on $S$, the dynamics of the system is
conditionally independent with the environment, characterizing the
phenomenon that the system only interacts within a niche of the
environment.  This phenomenon has also motivated the concept of {\it
Markov blanket} in graphical models \cite{10.5555/52121}, and in the
study of self-organization \cite{Palacios2020}: it is efficient,
perhaps only tractable, to estimate conditional probability measure
instead of the joint probability. Formally, let $\ca{O}_{0} = \{X\}$;
we create a binary random variable $H_{s}$ valued in $\{0, 1\}$, whose
probability measure $\mu_{s}(\bv{h}_s | \ca{O}_{s-1})$ is estimated by
solving an entropy maximization problem, referred as {\bf enclosed
maximum entropy problem}, given as follows---we shall explain the
subscript $s$ and $\ca{O}_{s-1}$ shortly after, and here $s$ could be
taken as $1$.
  \begin{displaymath}
    \max  -\sum_{\bv{h}_s \in \bb{B}^{n_s}}\mu_{s}(\bv{h}_s | \ca{O}_{s-1})\log \mu_{s}(\bv{h}_s | \ca{O}_{s-1})
  \end{displaymath}
  subject to
  \begin{align}
    \sum_{\bv{h}_s \in \bb{B}^{n_s}}\mu_{s}(\bv{h}_s | \ca{O}_{s-1})                   & = 1 \nonumber                              \\
    \bb{E}_{\mu_{s}(\bv{h}_s | \ca{O}_{s-1})}[\prod_{s'=1}^{s}h^{s'}_{i_{s'}}x_{i_0}] & = \tau^{s}_{\bv{i}}, \forall \bv{i} \in \otimes_{s'=0}^{s} [n_{s'}]\label{eq:coupling_equality-pre},
  \end{align}
  where $\{\tau^s\}_{\bv{i} \in \otimes_{s'=0}^{s} [n_{s'}]}, \tau^s \in \bb{R}$
parameterizes the coupling among random variables, $\otimes$ denotes
Cartesian product, and $n_{s'}$ denotes the dimension of $X$ (when
$s'=0$) and $H_{s'}$ (when $s' > 0$). For example, $\{X^{(j)}\}_{j \in I
\subset [m]}$ could be variants of edges of a certain orientation, and for a
given $i_1$, only in this set $H^1_{i_1} = 1$. Then, the left side
of \cref{eq:coupling_equality-pre} computes the average of the set
$\{X^{(j)}\}_{j \in I}$ (e.g., the average ``shape'' of the edges), that
characterizes the statistical coupling between $X_{i_0}$ and
$H^1_{i_1}$. Thus, $H_s$ is a coarse-grained variable that represents
a group of behaviors/events encoded by random variable $X$, and we
refer them as \textbf{object random variables}; and $H_{s}$ is
specific decisions (action of neurons) that represent certain objects
are detected.

\item The boundaries and niches emerge hierarchically: a great number
of neurons compose a neural system where the signals are the
activations of the sensory neurons, and the actions are the
activations of the motor neurons; the neural system occupies a niche
in a biotic body; and the boundary is formed by the neurons at the
exterior of the system that exchange signals with the
environment. Therefore, the modules at the higher part of the
hierarchy of the system receive and process events at a coarser
granularity that represent certain collective behaviors of a group of
events at the lower part of the hierarchy of finer granularity. To
appreciate the hypothesis, we might relate to the {\it binding
phenomenon} in human perception: human perceives the world not as the
interaction of pixels, but as the interaction of objects; that is, the
visual system binds elementary features into objects
\cite{Eguchi2018}. And thus human perception seems to operate on the
granularity of event groups. Therefore, the enclosed maximum entropy
problem previously is a hierarchy of optimization problems indexed by
integers $s \in [S], S \in \bb{N}^{+}$, which we refer as the
\textit{scale} parameter, and the \cref{eq:coupling_equality-pre} at
scale $s$ characterizes hierarchical coupling among events below scale
$s$, and $\ca{O}_{s-1} := \{X\} \cup \{ H_{i}\}_{i \in [s-1]}$. Notice that
a $H_s$ at higher scales is coupled with and coarse-grains over an
exponential number of states of object random variables at low scales.

\item Meanwhile, the signals from the outermost exterior of the system
are actually from the environment, and typically have fitness
consequences; for example, the photons observed by the
photon-sensitive cells in the eyes could come from nearby
plants. These signals are typically coarse-grained events
\cite{Flack2017a}: plants are of a large number of species, different
morphogenesis stages (determining whether a follower is mature enough
to have nectar), and etc. In response to these signals, the system
needs to act to reduce the uncertainty in predicting the development
of these coarse-grained events with fitness consequences. This is the
internal dynamics of the outermost exterior boundary that encloses the
whole system, and which as discussed previously, is formed by
hierarchy of boundaries and niches within the system. This hierarchy
of subsystems collectively estimate a conditional measure
$\mu(y | \bv{x})$---where the label $Y$ could be understood as
coarse-grained variable with fitness consequences and is also the
object random variable $H_{S}$ at a top layer---through estimating a
hierarchy of conditional measures $\{\mu_s\}, s \in [S]$ that characterize
the co-occurrence of the observed coarse-grain events (i.e., events at
relatively high scales), and the group of hierarchical events that
compose the coarse-grained events.  An Umwelt is formally (formal
definition given in \supps B B 4) a recursively extended probability
space that supports all the random variables $\ca{O}_S$ previously,
and could be understood as a model of the environment encoding a
hierarchy of events with fitness consequences through probabilities.
\end{enumerate}

As an ending note, now we could see connection between Boltzmann
distribution and Umwelt: compared with the disorganized complexity in
statistical physical, where the system maximizes entropy subject to
the constraint that system's energy (i.e., the average of energy of
units) is of a certain value, and the Umwelt organized-complex system
maximizes entropy subject to hierarchical event coupling that encodes
regularities in the environment.

\subsection{Stochastic, or Bayesian-probabilistic-graphical, definition of DNNs}
\label{sec:bayes-defin-dnns}

Though the Umwelt system introduced in \cref{sec:umwelt} is motivated
from physics, biology and complexity science, the design of the system
is guided by the goal that it should give a {\it whitebox} definition
of DNNs. Actually, a DNN is a Umwelt, and this observation leads to a
stochastic definition of DNNs that is a multi-layer probabilistic
graphical model and is a hybrid of Markov random field and Bayesian
network. This stochastic definition of DNNs is the foundation of the
analysis in rest of this work. We introduce the stochastic definition
of DNNs conceptually as follows, and a more rigorous formalism,
detailed derivations and discussion on Bayesian aspects are given in
\supps B B and \supps B C.

The enclosed maximum entropy problem is a classic statistical problem
whose solution belongs to the well known exponential families
\cite{wainwright2008graphical}.  The solution to the enclosed maximum
entropy problem is of the following parametric form:
\begin{equation}
  \label{eq:NN_exponential_distribution}
  \mu_{s}(\bv{h}_s | \ca{O}_{s-1}) = \frac{1}{Z}e^{\sum_{\bv{i} \in \otimes_{s'=0}^{s} [n_{s'}]}\lambda^{s}_{\bv{i}}\prod_{s'=1}^{s}h^{s'}_{i_{s'}}x_{i_0}},
\end{equation}
where
\begin{equation}
  \label{eq:partition_function}
  Z = \sum_{\bv{h}_s \in \bb{B}^{n_s}}e^{\sum_{\bv{i} \in \otimes_{s'=0}^{s} [n_{s'}]}\lambda^{s}_{\bv{i}}\prod_{s'=1}^{s}h^{s'}_{i_{s'}}x_{i_0}},
  \lambda^s_{\bv{i}} \in \bb{R}.
\end{equation}
And the Lagrange multipliers $\{\lambda^{s}_{\bv{i}}\}$ are obtained by
maximizing the loglikelihood of the law $\mu(\bv{h}_s | \ca{O}_{s-1})$
of the object random variables. This hierarchy of optimization gives a
law of all object random variables as
\begin{displaymath}
  \mu(\ca{O}_{S}) = \prod_{s=1}^{S}\mu_s(\bv{h}_{s} | \ca{O}_{s-1}).
\end{displaymath}
Because the goal is only to compute a $\mu_{S}(h_{S} | \bv{x})$ that
estimates $\mu(y | \bv{x})$, only loglikelihood of $\mu_{S}(h_S | \bv{x})$
is maximized; that is, a probability measure is estimated such that it
would give maximal likelihood to the sample $S_m$. Thus, a practice is
that, for each scale $s$, object random variables in $H_s$ are
randomly initialized to randomly coarse-grain events from the previous
scales, which is implemented by randomly initializing the parameters
(i.e., Lagrange multipliers) $\lambda^s_{\bv{i}}$. Correspondingly, the
$\{\lambda^{s}_{\bv{i}}\}_{s < S}$ of lower scales are modified to maximize
$\mu_{S}(h_S| \bv{x})$ instead of being modified to maximize
$\mu_s(\bv{h}_s | \ca{O}_{s-1})$. Actually, there is no such
$\mu_s(\bv{h}_s | \ca{O}_{s-1})$ to maximize because no ground truth on
$\{H_s\}_{s < S}$ is available, and the optimization of
$\{\lambda^{s}_{\bv{i}}\}_{s < S}$ to maximize $\mu_{S}(\bv{h}_{S} | \bv{x})$
should be understood as a statistical inference that infers
hierarchical event coupling that best maximizes the likelihood of the
observed coarse-grained random variables $H_{S}$---any values of the
$\{\lambda^{s}_{\bv{i}}\}_{\bv{i} \in \otimes_{s'=0}^{s-1}[n_{s'}]}$ parameterize an
exponential-family distribution that satisfies a set of
hierarchical-coupled constraints.  At a high level, the maximization
is implemented as an iterative algorithm: first, the marginal
$\mu_{S}(h_S | \bv{x})$ (i.e., likelihood estimated from the current
parameters) is computed, and maximized; then another iteration
repeats, if the optimization has not converged.

However, the computation of $\mu_S(h_S | \bv{x})$ is intractable, and
thus approximation is needed. A particular approximation scheme would
be DNNs. At each scale, the parameters $\lambda^s_{\bv{i}}$ of the
probability kernels $\mu_{s}(\bv{h}_{s} | \ca{O}_{s-1})$ are tensors. We
reparameterize the tensors into products of scalars as follows.
\begin{equation}
  \label{eq:Lagrange_reparameterization}
  \lambda^{s}_{\bv{i}} = \prod_{s=1}^{s}w^{s}_{i_{s-1}i_{s}},
\end{equation}
where $w^{s}_{i_{s-1}i_{s}} \in \bb{R}, i_{s-1} \in [n_{s-1}], i_{s} \in
[n_{s}]$.  Note that parameters are reused in the reparameterization
given in \cref{eq:Lagrange_reparameterization}. For example,
\begin{displaymath}
  \lambda^{s+1}_{\bv{i}} = w_{i_{s} i_{s+1}}\lambda^{s}_{\bv{i}_{:s+1}}.
\end{displaymath}
Thus, $\lambda^{s+1}_{\bv{i}}$ are parameterized in reference to
$\lambda^{s}_{\bv{i}}$.  We refer the parameterization as {\bf hierarchical
parameterization}.  As a result, $\mu_{s}(\bv{h}_s | \ca{O}_{s-1})$ in
\cref{eq:NN_exponential_distribution} is reparameterized as
\begin{equation}
  \label{eq:s-system-scalar-form}
  \mu_{s}(\bv{h}_s | \ca{O}_{s-1})
  = \frac{1}{Z}e^{\sum_{\bv{i} \in \otimes_{s'=0}^{s} [n_{s'}]}\prod_{s'=1}^{s}h^{s'}_{i_{s'}}w^{s}_{i_{s'-1} i_{s'}}x_{i_0}}.
\end{equation}
Collecting the scalars into matrices, we have
\begin{equation}
  \label{eq:s-system-matrix-form-pre}
  \mu_{s}(\bv{h}_s | \ca{O}_{s-1}) = \frac{1}{Z}e^{\bv{x}^{T}\overrightarrow{\Pi}_{s'=1}^{s}\bv{W}_{s'}\dg(\bv{h}_{s'})}.
\end{equation}
This is already a DNN in energy-based learning form. Thus, from now
on, we change the scale index $s$ to layer index $l$, and we call
those object random variables ($\{H_{l}\}_{l \in [L]}$) \textbf{neuronal
gates}.

Still, even in such dynamic programming parameterization, the
computation of marginals $\mu_{l}(\bv{h}_l | \ca{O}_{l-1})$ is still
intractable, and in the following, an approximate inference is
discussed, whose degenerated case is the well known ReLU activation
function \cite{Glorot2011}: actually, ReLU could be understood as an
entangled transformation that combines two operations into one that
computes an approximation of the $\mu_{l}(\bv{h}_{l} | \ca{O}_{l-1})$:
first, a Monte Carlo sample of $H_l$ is made; then the logit (i.e.,
$\bv{x}^{T}\overrightarrow{\Pi}_{l'=1}^{l}\bv{W}_{l'}\dg(\bv{h}_{l'})$
in \cref{eq:s-system-matrix-form-pre}) of $\mu_{l}(\bv{h}_{l} |
\ca{O}_{l-1})$ is computed. First, we motivate the approximate
inference of the marginal of $H_l$ given a datum $\bv{x}$. Note that
$\mu_{l}(\bv{h}_{l} | \ca{O}_{l-1})$ is a measure conditioned on
$\ca{O}_{l-1}$. To compute $\mu_{l}$, random variables in $\ca{O}_{l-1}$
need to be observed. Recall that $\ca{O}_{l-1} := \{H_{l'}\}_{l' \in
[l-1]} \cup \{X\}$. Let us assume that $\{H_{l'}\}_{l' \in [l-2]} \cup \{X\}$
are observed. As a consequence, $\mu_{l-1}(\bv{h}_{l-1} | \ca{O}_{l-2})$
is known. Ideally, we would like to compute the marginal of $H_l$ by
the following weighted average
\begin{displaymath}
  \mu_{l}(\bv{h}_l | \bv{x}) = \sum_{\bv{h}_{l-1} \in \bb{B}^{n_{l-1}}}\mu_{l}(\bv{h}_l | \bv{h}_{l-1}, \ca{O}_{l-2})\mu_{l-1}(\bv{h}_{l-1} | \ca{O}_{l-2}).
\end{displaymath}
However, the computation involves $2^{n_{l-1}}$ terms, and is
intractable. To approximate the weighted average, we sample a sample
of $H_{l-1}$ from $\mu_{l-1}$. When the weight matrices $\bv{W}_{l-1}$
are large in absolute value, the sampling would degenerate into a
deterministic behavior. More specifically, let $\bv{W}_{l-1} :=
\frac{1}{T}\hat{\bv{W}}_{l-1}$, where $T \in \bb{R}$, and
$\hat{\bv{W}}_{l-1}$ is a matrix whose norm (e.g., Frobenius norm) is
$1$. When $T \rightarrow 0$, the sampled $H^l_{i}$ would simply be determined by
the sign of
$\bv{x}^{T}\overrightarrow{\Pi}_{l'=1}^{l-1}\bv{W}_{l'}\dg(\bv{h}_{l'})\bv{W}_{l}$---$T$
could be appreciated as a temperature parameter that characterizes the
noise of the inference. Let $\bv{x}_{l-1} :=
\bv{x}^{T}\overrightarrow{\Pi}_{l'=1}^{l-1}\bv{W}_{l'}\dg(\bv{h}_{l'})$,
then we have
\begin{equation}
  \label{eq:degenerated_H-pre}
  \bv{h}^{l}_{i} :=
  \begin{cases}
    1, &\text{ if $(\bv{W}^{T}_l\bv{x}_{l-1})_{i} > 0$}\\
    0, &\text{ otherwise}.
    \end{cases}
\end{equation}
Correspondingly, $\mu_{l-1}(\bv{h}_{l-1} | \ca{O}_{l-2})$ becomes a
Dirac delta function on a certain $\hat{\bv{h}}_{l-1}$ given by
\cref{eq:degenerated_H-pre}. Correspondingly, marginal $\mu(\bv{h}_l |
\bv{x})$ of $H_l$ given $\bv{x}$ degenerated into $\mu(\bv{h}_l |
\hat{\bv{h}}_{l-1}, \ca{O}_{l-2})$, which is
\begin{equation}
  \label{eq:approximate_inference_mu_s-pre}
  \frac{1}{Z}e^{\bv{x}^{T}\overrightarrow{\Pi}_{l'=1}^{l}\bv{W}_{l'}\dg(\bv{h}_{l'})}.
\end{equation}
The previous computation is exactly what ReLU does. To see it more
clearly, we rewrite ReLU as the product of the realization of object
random variable $H_l$ and $\bv{W}^{T}_l\bv{x}_{l-1}$,
\begin{displaymath}
  \text{ReLU}(\bv{W}^{T}_l\bv{x}_{l-1}) := \dg(\bv{h}_l)\bv{W}^{T}_l\bv{x}_{l-1}.
\end{displaymath}
ReLU first computes a Monte Carlo sample of $H_l$, and then computes
the logit of \cref{eq:approximate_inference_mu_s-pre}. Therefore, ReLU
first computes an approximate inference of $\mu_l(\bv{h}_{l} |
\ca{O}_{l-1})$, and then computes to prepare for the inference of
$\mu_{l+1}(\bv{h}_{l+1} | \ca{O}_l)$ at next layer.

Therefore, the optimization of DNNs implements a classic {\it
generalized expectation maximization} algorithm \cite[section
9.4]{Bishop:2006:PRM:1162264} that maximizes the loglikelihood that
involves latent variables and is intractable to compute exactly. More
specifically, first, the approximation inference (implemented by
forward propagation) maximizes $\mu_l(\bv{h}_l | \ca{O}_{l-1})$ by
estimating realizations of $H_{l}$ while keeping the parameters
$\bv{W}_l$ fixed. And in the degenerated low temperature setting, the
Monte Carlo sample is the expectation of $H_{l}$. The forward
propagation maximizes a lower bound of $\mu_{L}(h_{L} | \bv{x})$. Then,
the approximated loglikelihood of $\mu_{L}(h_{L} | \ca{O}_{L-1})$ is
increased by modifying all weights $\{\bv{W}_l\}_{l \in [L]}$ through
gradient descent, while keeping $\{\bv{h}_{l}\}_{l \in [L]}$ fixed. This
step decreases the KL divergence between $\mu_{L}(h_{L} | \ca{O}_{L-1})$
and $\mu(y | \bv{x})$. Afterwards, the iteration starts again until a
local minimum is reached. Overall, this optimization optimizes for
hierarchical coupling of events represented by $\{H_l\}_{l \in [L]}$
that maximizes the loglikelihood of $\mu_{S}$ upon the sample $S_m$.

To conclude, the definition emphasizes an {\it ensemble}---a more
appropriate word here would be statistical, but ``statistical'' might
not ring a bell, and thus we intentionally use a less
institutionalized word ``ensemble'' here---perspective of DNNs. In this
ensemble perspective, a DNN is {\it not} viewed from the perspective
of functional analysis as a {\it function} that takes a datum and
returns an output, but as a {\it statistical-inference system} that
infers the regularities in the environment through the hierarchical
coupling among the neuron ensemble, and the coupling between the
neuron ensemble and the environment (a sample of data)---recall that
``statistics'' means to infer regularities in a population.

\subsection{Self-organization of DNNs}
\label{sec:dnn-self-organ}

\subsubsection{Ising model and self-organization}
\label{sec:ising-model-self}

The atomic/mechanistic view of materials proposed by L. Boltzmann did
not prevail until the statistical mechanical model was developed for
materials that are not just gases, but also liquid and solids
\cite{Brush1976}. And this development revolved around the classic
{\bf Ising model}. Unlike freely moving units in gases, the units in
an Ising model are positioned on lattices with specific coupling
forces among the units' neighboring units, and the collective
behaviors of the units are characterized by a Boltzmann distribution
whose energy function characterizing stronger coupling force than that
of units in gases. Ising model abstracts away the intricate quantum
mechanism underlying the units and their interaction, and simply
characterizes them as random variables taking discrete values in
$\{+1, -1\}$. As in the case of the Boltzmann distribution, by the
time Ising model was proposed, it was not clear at all that such a
simplistic model could characterize the collective behaviors (i.e.,
phase transitions, which we shall discuss in
\cref{sec:plast-phase-benign}) of liquid or solids. The clarity
conveyed by the Ising model resulted from arduous efforts that spanned
about half a century \cite{Brush1976}.

From the perspective of history of science, the study of DNNs has the
same problem identifying a simple-but-not-simpler formalism that could
analyze the collective behaviors of neurons in experiments. To
introduce such a formalism of DNNs, it is informative to put DNNs and
statistical-physical systems (e.g., magnets) under the umbrella
concept of {\it self-organization}, which we describe as follows.

The behaviors of physical systems characterized by Ising models, and
these of DNNs both are behaviors referred as self-organization
\cite{Ashby1947,nicolis1977self,VonFoerster2003,Heylighen2002,10.5555/1212651,camazine2003self,DeWolf2005}.
Self-organization refers to a dynamical process that an open system
acquires an organization from a disordered state through its
interaction with the environment, and such an organization is acquired
through distributed interaction among units of the system---and thus
from an observer's perspective, the system appears to be
self-organizing---we note that it is still an open problem
\cite{Keller2008,Keller2009} to unambiguously define a self-organizing
process
\cite{Spitzner1998,Gershenson2003,Shalizi2004,Polani2008,Prokopenko2009,Rosas2018,Gershenson2020},
and the dynamical-system formalism is among more established
definitions of its formal definitions
\cite{Ashby1947,VonFoerster2003}.  For example, when a ferromagnetic
material is quenched from a high temperature to a temperature below
the critical temperature of the material under external magnetic
field, the material acquires an organization where the spin direction
of molecules in the material spontaneously align with the external
field, and at the same time, the system also maintains a degree of
disorder by maximizing the entropy of the system (for example, the
molecules vibrates and exchanges location with one another
rapidly). This process is similar to the learning process of a
Umwelt/DNN: in the process, the system acquires an organization by
estimating a probability measure that could, for example, predicts the
probability of an example being a face of particular person, and at
the same time, maintains a degree of disorder to accommodate the
uncertainty of unobserved examples/environment by maximizing the
entropy of the system.

The self-organization of magnets results from, or is abstracted as,
the minimization of a potential function referred as {\it free
energy}, and is analyzed by the Ising model. Next, building on the
stochastic definition of DNNs given previously, we shall present a
formalism that characterizes the self-organizing process of DNNs and
enables the analysis of the coupling among neurons---a more rigorous
formalism and more details are given in \supps B D.

\subsubsection{Self-organization of DNNs through a feedback-control loop composed
  of coarse-grained variable and hierarchical circuits}
\label{sec:self-organ-dnns}

\begin{figure*}[t]
  \centering
  \includegraphics[width=\linewidth]{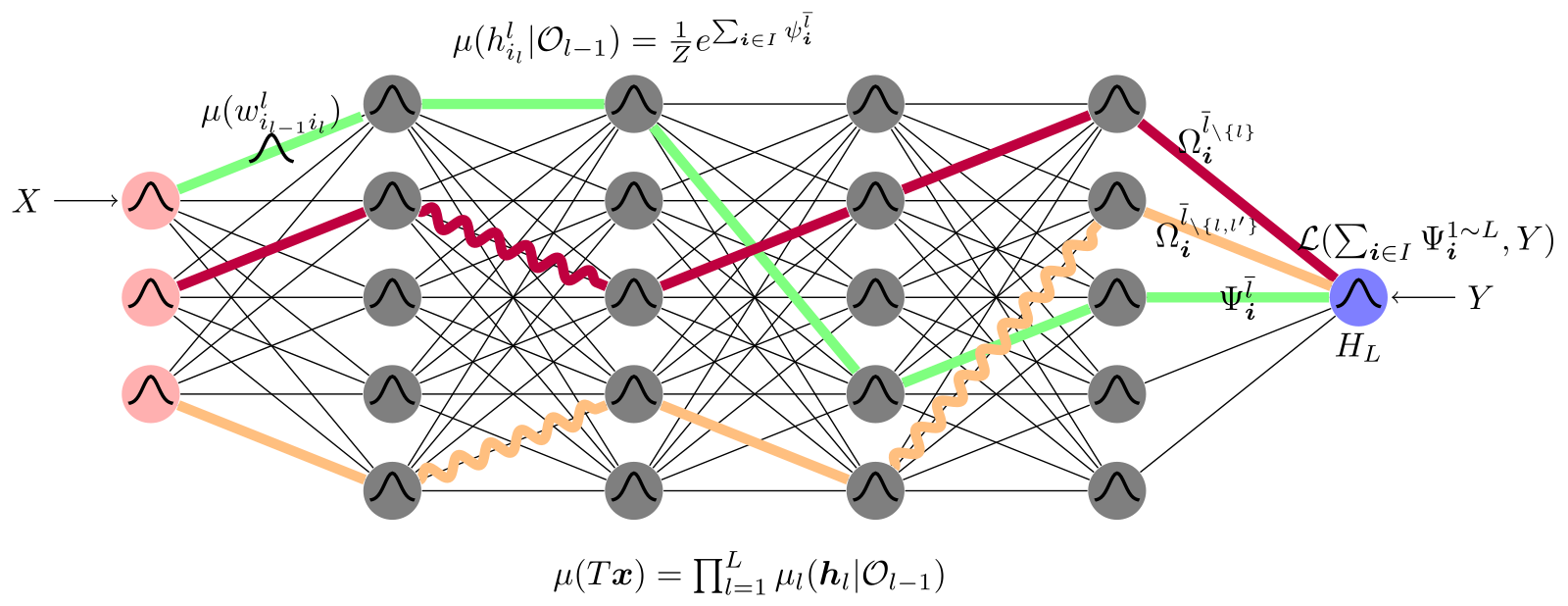}
  \caption{Illustration of the multilayer probabilistic graphical
model, the formalism of circuits and the feedback-control loop of
DNNs. The learning of Umwelt is through a feedback-control loop
composed by hierarchical circuits and the coarse-grained variable
computed by the circuits. The loop is graphically illustrated as
follows. $l$ denotes layer index. Each node represents a neuronal
gate, denoted by $H^{l}_{i_l}$ and each edge represents a weight
connecting between two neuron gates, denoted by $W^{l}_{i_{l-1}
i_l}$---neuronal gate is a concept introduced in
\cref{sec:bayes-defin-dnns}, and replaces the activation function in
deterministic DNNs with a statistical inference on a random variable
supported on $\{0, 1\}$. A neuron constituents a gate and the weights
connected to the gate. All neuronal gates and weights are random
variables---this is emphasized by drawing a bell-shape curve inside each
gate, and on the weight at the upper left corner. A path connected
gates through weights from left to right (not necessarily the
rightmost) is a basis circuit, denoted by $\Psi^{\over{l}}_{\bv{i}}$,
that represents the multiplication of all connected random variables,
both gates and weights, illustrated as the thick green path;
$\overl{l}$ denotes the tuple $(1\ldots l)$, and $1\sim L$ denotes the tuple
$(1\ldots L)$---those tuples index basis circuits. The basis circuits with
curvy edges (i.e., $\Omega^{\overl{l}_{\setminus \{l\}}}_{\bv{i}}$,
$\Omega^{\overl{l}_{\setminus \{l, l'\}}}_{\bv{i}}$) represent perturbation of
basis circuits induced by perturbations of weights (represented by
curvy lines), which are simply basis circuits with the perturbed
weights being differentiated out. All basis circuits end with a same
gate are arithmetically added together, and are referred as a neuron
assembly.  Each neuron solves a maximum entropy problem, and thus
conditioning on gates in the previous layers, is of an exponential
family distribution given by $\mu(h^{l}_{i_l} | \ca{O}_{l-1})$ at the
top, where $\ca{O}_{l-1} := \{X\} \cup \{ H_{i}\}_{i \in [l-1]}$ and the
energy function is the realization of addition of all basis circuit
$\Psi^{\overl{l}}_{\bv{i}}$ connected to it---therefore, all connected
gates are statistically dependent with one another. The weights are
samples of a prior distribution at the beginning of training. The
network $TX$ (i.e., all the weights and gates) is a stochastic system
whose law is $\mu(T\bv{x})$ given at the bottom, which factorizes into
conditional measures $\mu(\bv{h}_l | \ca{O}_l)$ of neurons by
layers. The addition coarse-grains over the output of all the basis
circuits, and thus computes a coarse-grained random variable; for
example, the blue rightmost gate $H_L$ is the output of the DNN that
estimates the class of the example $X$ perceived by the red leftmost
neurons.  For each realization (example) of the perceptual signals
$X$, the network computes a realization of the coarse-grained variable
$H_L$. The feedback signal (i.e., labels) $Y$ from the environment
(i.e., dataset) is compared against the realization of $H_L$ through a
surrogate loss function $\ca{L}(\sum_{\bv{i} \in I}\Psi^{1\sim L}_{\bv{i}}, Y)$,
and the discrepancy is back-propagated back to adjust the weights to
control the errors. This forward-backward propagation composes one
iteration of the loop.}
  \label{fig:dnn_graph}
\end{figure*}

We present a circuit formalism that characterizes the risk
minimization of DNNs as a self-organizing informational-physical
process where the system self-organizes to minimize a potential
function referred as {\it variational free energy}
\cite{Friston2006,Friston2009,Friston2010,Pezzulo2018,Buckley2017,Yufik2016,Yufik2017}
that approximates the entropy of the environment by executing a {\it
feedback-control loop}
\citep{Rosenblueth1943,10.5555/1096919,Heylighen2001} that is composed
by the hierarchical {\it circuits} (implemented by coupling among
neurons) and a {\it coarse-grained random variable} computed by the
circuits. The feedback-control loop shall provide the formalism to
study the symmetries in DNNs.

Recall that for physical systems, without energy intake, any processes
minimize free energy, and free energy could be decomposed into the
summation of averaged energy and negative entropy as
\begin{equation}
  \label{eq:free_energy_decom}
  F = \bb{E}_{\mu}[E(\bv{x})] - T S_{\mu},
\end{equation}
where with an abuse of notation, $T$ here denotes temperature instead
of a DNN. And specifically for the self-organizing process of a
physical system, the process dissipates energy and reduces entropy
\cite{Prigogine1978}. A DNN is a computational simulation and a
phenomenological model of biotic nervous system, and thus it could be
appreciated as a model that characterize the informational perspective
of the biotic system, and ignores the physical perspective.

More specifically, the maximization of the hierarchical entropy
previously approximates the entropy (i.e., $S_{\mu}$) of the environment
(dataset). Recall that a loss function in statistical learning
\cite{Vapnik1998} is a surrogate function that characterizes the
discrepancy between the true measure $\mu(y | \bv{x})$ and the
approximating measure $\mu(h_{L} | \bv{x})$. With a logistics risk
(i.e., cross entropy), we would have the risk given as
\begin{displaymath}
    S_{\mu(y | \bv{x}), \mu(h_{L} | \bv{x})} := -\bb{E}_{(\bv{x}, y) \sim \mu(y | \bv{x})}\left[ \log \mu(h_L | \bv{x}) \right],
\end{displaymath}
which is maximization of negative loglikelihood discussed in
\cref{sec:bayes-defin-dnns}. Further notice that, cross entropy could
be decomposed into the summation of the KL divergence $D_{\KL}(\mu ||
\nu)$ and entropy $S_{\mu}$ as
\begin{displaymath}
  S_{\mu(y | \bv{x}), \mu(h_{L} | \bv{x})} = S_{\mu(y | \bv{x})} + D_{\KL}(\mu(y | \bv{x}) || \mu(h_{L} | \bv{x})).
\end{displaymath}
Thus, the logistics risk is an upper bound on the KL divergence
between $\mu(y | \bv{x})$ and $\mu(h_{L} | \bv{x})$ ($S_{\mu(y | \bv{x})} \geq
0$): suppose that at a minimum, KL divergence becomes zero, the cross
entropy is the entropy of $\mu(y | \bv{x})$; that is, the entropy of the
dataset (environment).

Therefore, the hierarchical maximum entropy problem optimizes the
Lagrange multipliers for a variational approximation to the entropy
part of the free energy of the environment, and thus could be
appreciated as minimizing variational free energy, although the
message has been formulated in a rather convoluted way initially under
a different context \cite{Friston2006}.

To analyze this variation free energy minimization process, we develop
a formalism of feedback-control loop. However, to begin with, we
discuss some background.
\begin{enumerate}[leftmargin=0cm]
\item The macroscopic behaviors---that is, the order at a higher
scale---of a self-organizing system are the coarse-grained effect of
symmetric microscopic units in the system; for example, for magnetic
systems, it is the macroscopically synchronized symmetries of
microscopic spins that are perceived as, e.g., magnetic force.  Thus,
the system could be studied through interaction between the symmetries
and the macroscopic behaviors of the system, by typically formulating
variables that coarse-grain over the symmetric behaviors of
constituent units in the Ising model (through statistical field
theory, e.g., techniques such as renormalization groups
\cite{kadanoff2000statistical,McKay1982,ROSATI2001,Caglar2017}).
\item However, unlike the {\it homogeneity} in physical systems, a
biotic system consists of a vast number of {\it heterogeneous}
microscopic units that interact locally in a {\it hierarchically
coupled} way against goal-directed feedback signals and induce an
emergent macroscopic phenomenon.  The spatially and temporally
heterogeneous units, and the coupled interaction among units results
in the phenomenon that a microscopic change at one scale has
implications ramifying across scales throughout the system
\cite{Levin1997,Krakauer2011}. Therefore, in biotic systems, the
coarse-grained variables are not coarse-grained characterization of a
group of units whose higher order coupling could be considered as
irrelevant fluctuations and be safely discarded (e.g., field in
physics
\cite{kadanoff2000statistical,McKay1982,ROSATI2001,Caglar2017}), but
are representation of the environment computed by the system itself to
capture coarse-grained regularities.
\item Such coupling in biotic systems has been studied through
analyzing the feedback-control loop between hierarchical circuits (or
simply circuits) in the system and the coarse-grained variables
computed by the circuits \cite{Mossio2016,Flack2017a}. The circuits
are networks composed by the coupled heterogeneous units that intake
feedback (signals from the environment and other units in the system),
perform internal computation, and effect actions, and meanwhile
compute the coarse-grained variables in the process.  For example, in
the genotype of an organism, the hypothesis of the organism is encoded
as the coupling of heterogeneous genes, and could temporally and
spatially regulate the development of the organism. The regulatory
process could be studied by analyzing the hierarchical circuits (e.g.,
gene regulatory networks) that implement feedback-control loops with
hierarchically coupled regulatory logic operation and gene expression
(which for example, could generate regulatory signals)
\cite{Davidson2006,Davidson2010}---hierarchy here refers to, for
example, the animal body development determined by gene regulatory
networks, where each phase of the development encoded by the genes has
beginning, middle and terminal stages, and later events recursively
refine the body parts developed in early events \cite{Davidson2006}. A
more detailed introduction to the methodology in theoretical biology
is given in \supps D B.
\end{enumerate}

Thus, first we develop a circuit formalism. The exponent in
\cref{eq:s-system-matrix-form-pre} could be rewritten as
\begin{equation}
  \label{eq:assembly-pre}
   \sum_{\bv{i} \in I}X_{i_0}\Psi^{\wedge 1\sim L\wedge }_{\bv{i}},
\end{equation}
  where
\begin{equation}
  \label{eq:basis-circuit-pre}
  \Psi^{\wedge 1\ldots L \wedge}_{\bv{i}} := \prod_{l=1}^{L-1}W^{l}_{i_{l-1} i_{l}}H^{l}_{i_{l}}W^{L}_{i_{L-1} i_{L}},
  I = \overrightarrow{\otimes}_{l\in \bb{L}}[n_{l}],
\end{equation}
and we have used upper case notations instead of lower case because
all the variables there are random variables---the exponent is the
realization of those random variables:
\cref{eq:s-system-matrix-form-pre} is the law of gates conditioned on
weights and gates of previous layers, and the law of weight matrices
are symmetric at initialization, but would change after training
starts.  We call $\Psi^{\wedge 1\ldots L \wedge}$ a \textbf{basis circuit}, and the
wedge symbol is also part of the formalism that denotes the circuit is
missing an input or an output---$X_{i_0}$ is written as the input
explicitly and thus $\Psi^{\wedge 1\ldots L \wedge}_{\bv{i}}$ has a wedge symbol on the
left. For convenience, we also write $\Psi^{\wedge 1\ldots L \wedge}_{\bv{i}}$ as $\Psi^{\wedge
1\sim L \wedge}_{\bv{i}}$. In addition, a basis circuit is defined
recursively, and thus a segment of basis circuit is also a basis
circuit: for example,
\begin{displaymath}
  \Psi^{\wedge 1\ldots l' \wedge}_{\bv{i}_{:l'+1}} := \prod_{l=1}^{l'-1}W^{l}_{i_{l-1} i_{l}}H^{l}_{i_{l}}W^{l'}_{i_{l'-1} i_{l'}},
\end{displaymath}
where $l' < L$.  Further notice that \cref{eq:assembly-pre} is the
coarse-graining/addition of a large number of basis circuits, and we
refer it as an \textbf{assembly}.

Under the circuit formalism, the risk minimization of a DNN
$T(X)$ could be written as
\begin{equation}
  \label{eq:feedback-loop-pre}
  \begin{aligned}[t]
  R & = \bb{E}_{(X,Y) \sim \mu}\left[ \max(0, 1 - YT(X) \right] \\
    & = \bb{E}_{(X,Y) \sim \mu}\left[ \max(0, 1 - Y\sum_{\bv{i} \in I}X_{i_0}\Psi^{\wedge 1\sim L\wedge }_{\bv{i}}) \right],
  \end{aligned}
\end{equation}
where we have used hinge loss as a concrete example to illustrate, and
in this case the output $T(X)$ of a DNN is simply the coarse-graining
of basis circuits---if logistics risk is used, the coarse-graining is
the logit and still needs to pass through a sigmoid function to
become the output.
Then, the dynamics of the self-organizing process of a DNN is a {\bf
feedback-control loop} given as the iterative process
\begin{equation}
  \label{eq:gd_differential_equation_pre}
  \frac{d \bv{\theta}}{d t} =  - \eta \nabla_{\bv{\theta}}R,
\end{equation}
where $\eta$ is a scalars that scale the gradient $\nabla_{\bv{\theta}}R$ and is
called as {\it step size}, or {\it learning rate} in the literature,
and $\bv{\theta}$ denotes all the trainable parameters of a DNN.  The loop
characterizes a dynamical process that executes a loop described as
follows: first, the circuit composed by neurons represents a
hypothesis that computes a prediction of a coarse-grained random
variable $H_L$; second, a surrogate risk that measures the discrepancy
between the prediction and the observed valued of the variable ($Y$,
feedback signals) from the environment/dataset; third, the circuit
self-organizes according to gradient/feedback back-propagated from the
feedback signals (the derivative of risk w.r.t. $T(X)$) at the top-layer
neuron; lastly, the process goes back to the first step. The preceding
concepts are illustrated in \cref{fig:dnn_graph}.

\subsection{Adaptive symmetries in the feedback-control loop}
\label{sec:adapt-symm-feedb}

Self-organization is a transdisciplinary concept that tries to
characterize both the physical and biotic systems
\cite{Keller2008,Keller2009}, and has the problem of ambiguous
definitions
\cite{Spitzner1998,Gershenson2003,Shalizi2004,Polani2008,Prokopenko2009,Rosas2018,Gershenson2020}.
The self-organization of magnets given in \cref{sec:dnn-self-organ} is
more precisely characterized as a {\it symmetry-breaking} process: as
the temperature decreases, the free energy of the system decreases,
and the system self-organizes and breaks symmetries known as the {\it
translation} and {\it rotation} symmetry; the breaking of rotation
symmetry leads to the alignment of the spins' spinning direction,
which collectively manifests as the magnetic field; and the breaking
of translational symmetry results in the phenomenon that magnetic
fields with varied strength exist at different spatial locations of
the system. Collectively, the breaking of the two symmetries is known
as {\it replica symmetry breaking} \cite[Chp~3]{nishimori01}. To
clarify, we have described the symmetry breaking of a particular type
of magnets known as spin glasses, and the spin glasses would be the
example to illustrate concepts from now on because some extraordinary
similarity exists between DNNs and spin glasses, which we would
intermittently discuss throughout this work.  Perhaps marvelously, the
self-organization of DNNs could also be characterized as a
symmetry-breaking process, and in the following, we characterize a
symmetry that we refer as {\bf circuit symmetry}.

\subsubsection{From conservative-symmetry in physics to adaptive-symmetry in
  biology}
\label{sec:from-cons-symm}

To begin with, we introduce the foundational role that symmetries play
in physics.  The symmetries in physics are formalized as symmetry
groups in mathematics. Symmetry groups formalize invariants of
physical systems that constituent the fundamental concepts to
understand these systems
\cite{bailly2011mathematics,Longo2011}:
\begin{enumerate}[leftmargin=0cm]
\item symmetries construct objectivity by identifying observables that
are stable at a spatial and time scale of human perception and thus
could be measured by instruments; such stable behaviors are
concentration-of-measure phenomena manifesting over time, and referred
as ergodic states \cite{Gallavotti2016} of the system that are related
by symmetric transformations that conserve the free energy;
\item the breaking of symmetries is associated with the change of a
system's stable behaviors, and thus characterizes the dynamics of the
system;
\item and the hypotheses derived from these mathematical invariant and
the experimentally validation of the hypotheses through measurements
on the observables constitute the fundamentals of the science of the
system.
\end{enumerate}
   For example, at a high temperature the rotation symmetry of the
spins are the stable invariant that characterizes the behaviors of the
system---rotation of the spins conserves the free energy. Energy
dissipation of the system decreases free energy, and thus breaks the
rotation symmetry. The breaking of the rotational symmetry
characterizes the dynamics of the system. The dynamical process can be
experimentally observed by measuring the magnetization, and the spin
glass order parameters. Therefore, symmetries formalizes the conserved
behaviors (e.g. free energy in a spin glass) of a physical system when
no external factors (e.g., energy) are influencing the system, in this
regards, symmetries in physical are {\bf conservative symmetries}. A
more detailed introduction is given in \supps D A.

\begin{figure*}[t]
  \centering
  \subfloat[\label{fig:weight_symmetry} Mixture of intact and broken
weight symmetries throughout DNN training. The figure has $12$
subplots that each plots the weights' distribution at a layer of the
DNN in the experiments throughout training---the DNN has $12$
layers. Each subplot has $10$ histograms plotted from the top to the
bottom, and plots the distributions at epoch $0, 10, 20, 30, 40, 100,
160, 200, 260, 300$, respectively. The black line at the center of
each ridgeline graph is the $y$-axis. The blue lines are the smoothed
histograms obtained by smoothing the frequency with a Gaussian
kernel.]{
    \includegraphics[width=\linewidth]{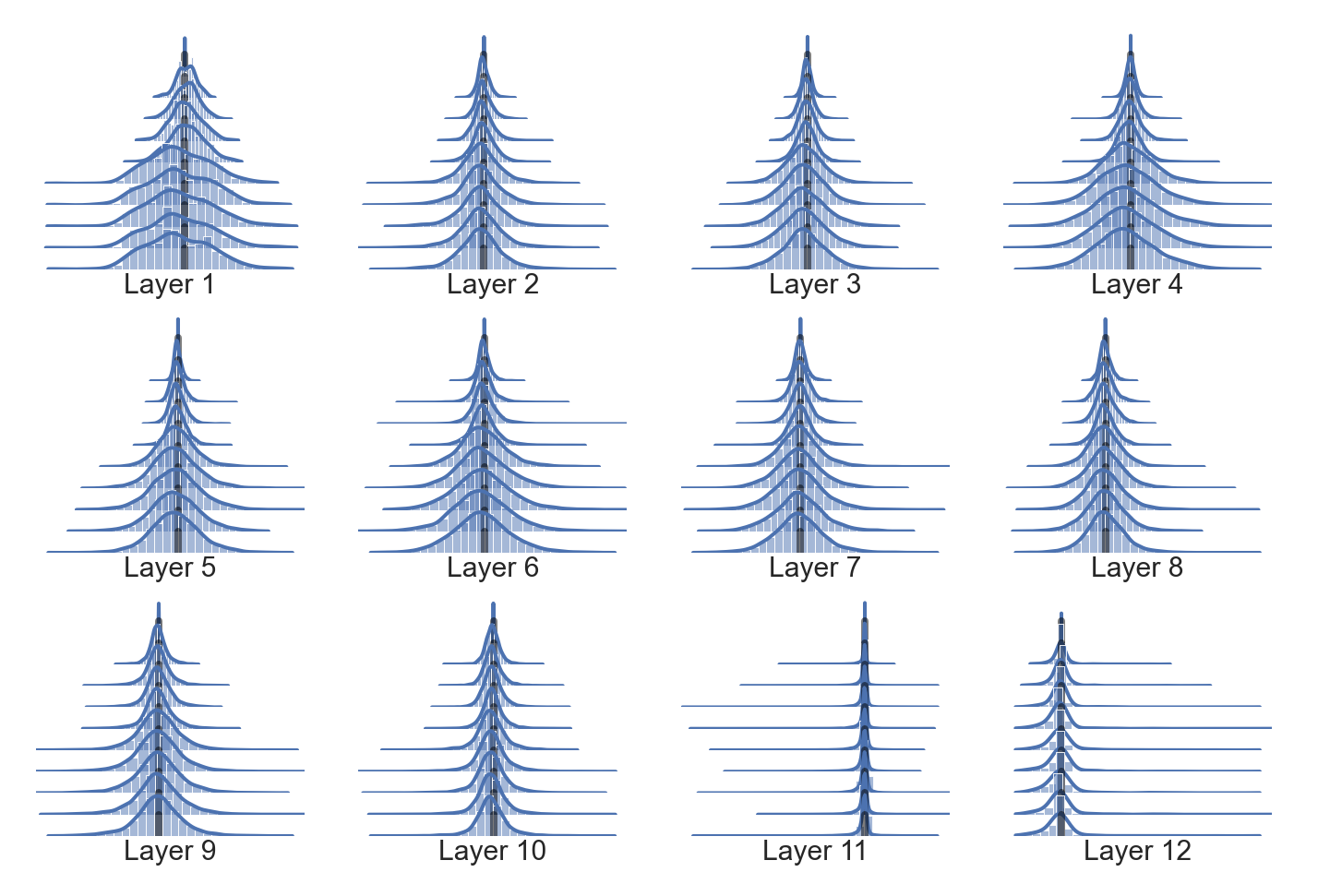}
  }

  \subfloat[\label{fig:circuit-symmetry} Circuit symmetries of basis
circuits, and perturbation of basis circuits of order $1, 2$ at the
beginning of DNN training. The three plots from left to right are the
histogram of basis circuits (which compute the output of a DNN),
histogram of perturbations of basis circuit of order $1$ (which
compute the gradient of a DNN), histogram of perturbation of basis
circuit of order $2$ (which compute the Hessian of a DNN). The basis
circuits are sampled from all possible basis circuits uniformly at
initialization. Though the number of circuits sampled is only a
varnishingly small fraction of all possibles circuits, the symmetry is
consistently observed for dozens of samples in experiments that are
not shown here.]{%
  \includegraphics[width=\linewidth]{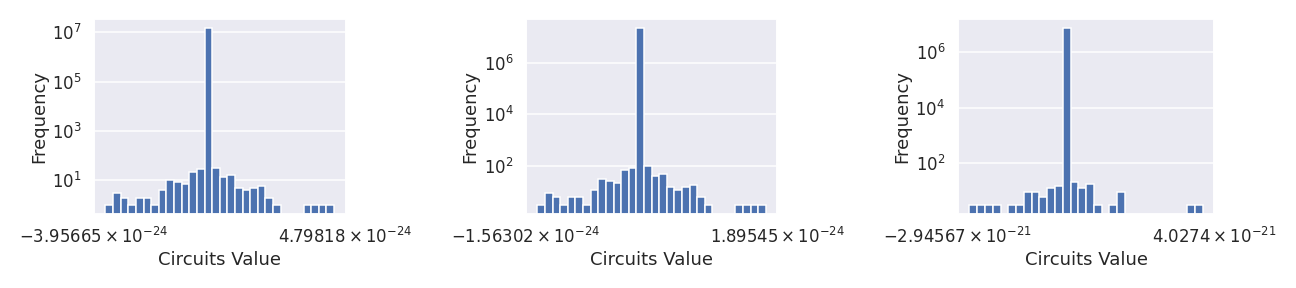}%
  }
  \caption{Microscopic adaptive symmetries in a DNN whose
specification is introduced in \cref{sec:problem-setting-intro}. Each
weight $W^{l}_{i_{l-1} i_l}$ (an edge in \cref{fig:dnn_graph}) is of a
law that is symmetric w.r.t. $y$-axis at the beginning of the
training, and fig. a shows that such weight symmetries only slight
skews throughout training, implying the majority of the weight
symmetries do not break during training. A basis circuit (a path in
\cref{fig:dnn_graph}) is the multiplication of neuronal gates and
weights, and thus the weak correlation among gates---as a result of the
fact that each gate is connected to a large number of neurons---and
weight symmetry are theoretically found (through a theorem) to induce
a composite symmetry of basis circuits, referred as the circuit
symmetry. Circuit symmetries at the beginning of the training are
experimentally shown in fig. b. A basis circuit of circuit symmetry is
of equal probability to contribute to the coarse-grained variable in
the feedback-control loop, positively (e.g., detecting a nectar-making
flower) or negatively; and in response to the feedback signals
(labels), circuit symmetries would break to decrease the errors (e.g.,
of detecting flowers).  The symmetries break heterogeneously: the
broken symmetries encode information about the environment/dataset,
and the intact symmetries maintain adaptability to reduce further
information perturbations (errors) of novel examples.}
\label{fig:micro-adaptive-symmetry}
\end{figure*}

However, in biotic systems, symmetries conserving free energy is
continually being broken and no symmetries of conservation exist that
constituent fundamental concepts to understand biotic systems;
instead, the invariant of variants, might be a fundamental concept
\cite{Longo2013,Montevil2016,Jost2021a}. In this work, we formalize a
stable invariant of variants as {\bf adaptive symmetry} or {\it
symmetry of adaptation}. We introduce the concept of adaptive symmetry
as follows.
\begin{enumerate}[leftmargin=0cm]
\item The symmetry is not like the symmetries of conservation in
physical systems that are induced by the preference of units to a
certain configuration to minimize free energy, but is a symmetry of
adaptation: the system has the capacity to process the novel
information---that is, to adapt---by posing in states where symmetric
possible directions to adapt could be adopted, which in turn is
induced by the complex cooperative interaction among the heterogeneous
units in a biotic system; and the symmetric states would break in
response to random fluctuations and external feedback signals
\cite{Li2010b,Flack2012}, and thus is also a typical self-organizing
process. The breaking of such symmetries results in functional
diversification on every scale, from molecular assembliers, to
subcellular structure, to cell types themselves, tissue architecture,
and embryonic body axes \cite{Li2010b}.
\item For example, the cell specification in the embryonic
differentiation could be conceptualized as a symmetry breaking
process: from a symmetric state where an embryonic cell has multiple
ways to adapt, in response to the feedback signals regulated
temporally and spatially by {\it gene regulatory networks}
\cite{Davidson2006,Davidson2010}, the cell breaks the symmetry, and
specifies into more specialized types of cells.
\item In the lifespan of a biotic system, the symmetries and broken
symmetries {\it coexist}, and might be a way to characterize
structural stability and adaptability of life \cite{Longo2011}: for
example, despite an organism was developed by breaking a sequence of
symmetries since the embryonic stage, immune cells could still break
symmetries and differentiate into specialized immune cells in response
to specific pathgens.
\end{enumerate}
     The study of biological symmetry breaking is still an on-going
scientific efforts; the relation between biological symmetry and
conservative symmetries in physics has not been fully understood, and
a formalism comparable to the Noether's theorem in physics is yet to
be formulated \cite{Longo2011,Longo2019}. We shall provide some
examples to speculatively discuss the interaction between conservation
symmetries in physics, and adaptive symmetries in biology when we
discuss the related works that apply conservative symmetries to DNNs
in \supps A C. And again, a more detailed discussion
is given in \supps D A.

\subsubsection{Circuit symmetry in DNNs}
\label{sec:circ-symm-dnns}

Each basis circuit (introduced in \ref{sec:dnn-self-organ}) is a
microscopic feedback-control loop that contributes to the computation
of the coarse-grained variable $H_L$. We identify an adaptive symmetry,
referred as {\it circuit symmetry}, of the basis circuits that
characterize the phenomenon that is of equal probability to contribute
to the coarse-grained variable positively or negatively---that is, for
example in binary classification problem, to contribute positively to
classify the pattern as $1$ (positive example), or negatively to
classify the pattern as $0$ (negative example). This circuit symmetry
would break to change the coarse-grained variable positively or
negatively, in response to the feedback signals, i.e., labels of the
data. We introduce the symmetry as follows, and a more rigorous
formalism and further details are given \supps B E.

A statistical-physical system is characterized statistically as an
ensemble of possible states where the system could be in. And
symmetries in physics characterize equivalent states in the sense of
free energy, and thus the probability of a system to stay in these
states. Thus, at equilibrium, symmetries in statistical physics
characterize the equivalence among ergodic states over a long-time
(compared with thermodynamic timescale).

In contrast, circuit symmetry also characterizes certain equivalence
among possible states realized from an ensemble of all possible
states, but there are no transitions among these states in the sense
of ergodicity---such a symmetry has been referred as {\bf stochastic
symmetries} in the context of complex networks
\cite[Chp~6]{wadhawan2011latent}. In addition, equivalence of circuit
symmetry is not in the sense of free energy, but of adaptability.

An adaptive symmetry, referred as {\bf weight symmetries}, exists in
DNNs; and the cooperative breaking of such symmetries would induce
system-wide ordered, or macroscopic, behaviors of DNNs. Operationally,
the weight symmetries simply characterize the phenomenon that each
weight of a DNN is sampled/realized from a random variable with a
symmetric law. Furthermore, note that each weight is realized from a
independent random variable. Such individualized symmetries imply that
the symmetries could be broken in a heterogeneous way---which we refer
as {\bf heterogeneous symmetries}---as in the biotic systems, where
macroscopic symmetries are composed by heterogeneous units whose
symmetries would break in response to the local, transient, even noisy
feedback signals each unit received, and whose symmetry breaking
cooperates to form stable system-level asymmetries \cite{Li2010b}.
For a large network, only a subset of weights' symmetry would break
throughout training to encode information, and statistically, the
weight symmetry still holds for the majority of the weights. This
could be seen in \cref{fig:weight_symmetry}, where we could see that
although the weight distributions gradually skew as training
progresses, they are still approximately symmetric w.r.t. $y$-axis
throughout the training.

Weight symmetries induce a {\bf composite symmetry} that are referred
as \textbf{circuit symmetry}.  Recall that feedback-control loop of
DNNs is composed by hierarchical basis circuits and the coarse-grained
variable, and the coarse-grained variable is computed by a neuron
assembly that is the addition of basis circuits. Thus, each (basis)
circuit in the loop is a \textit{microscopic feedback-control loop}
composed by neurons and the weights connecting
neurons. Correspondingly, weight symmetries induce the composite
circuit symmetry. Because weight symmetry is a heterogeneous symmetry,
and thus circuit symmetry is also a heterogeneous symmetry. As a
result, the circuits in a DNN could be of broken symmetry in only a
subset of all circuits, and the system is in a state where intact and
broken circuit symmetries coexist. This phenomenon is a
concentration-of-measure phenomenon and is characterized as a
probability bound in a theorem given in \supps B E 5. Informally, for any
basis circuits whose weights are of weight symmetry,
\begin{equation}
  \label{eq:circuit_symmetry_circuits_informal-pre}
      \mu(\psi^{1\sim L\wedge}_{\bv{i}}) \approx \mu(-\psi^{1\sim L\wedge}_{\bv{i}}).
\end{equation}
We also plot the histograms of basis circuits at the beginning of the
training in \cref{fig:circuit-symmetry} for demonstration: we only
plot the histograms at initialization because once the training
begins, the statistical behaviors of basis circuits that result from
random initialization are broken, and it is difficult to identify
basis circuits that are of circuit symmetry among the exponential
number of basis circuits without sophisticated efforts, which would be
a digression; instead, we simply uniformly sample basis circuits at
initialization for demonstration, and circuit symmetries could be seen
through such uniform sampling. The figure actually presents three
types of basis circuit, and in the current stage, we have not
explained in details the perturbations of basis circuits, which
compute derivatives of output of a DNN, and they could be understood
simply as basis circuits. As could be seen from
\cref{fig:circuit-symmetry}, the law of basis circuits are
symmetric---the output of a DNN, the gradient, the Hessian (i.e., neuron
assemblies), are simply the addition of the output of these basis
circuits, respectively.

\subsubsection{Statistical assembly methods}
\label{sec:stat-assembly-meth}

Recall that in \cref{sec:from-cons-symm}, we introduce that symmetries
in physics identify observables that characterize the
coarse-grained/macroscopic behaviors of the system. These
coarse-grained observables are typically self-similar to the
microscopic behaviors as a result of symmetries, and are calculated
through renormalization over symmetry groups---that is, the renowned
renormalization group technique. Further recall that the
coarse-grained variables (i.e., assemblies) computed by a DNN are the
addition basis circuits, and the addition is actually over symmetry
groups. A core of this work is to show that coarse-grained behaviors
of basis circuits that are self-similar to the behaviors of basis
circuits also exist for DNNs; in other words, the coarse-grained
effect of the microscopic adaptive symmetry would manifest as a
macroscopic/coarse-grained adaptive symmetry of DNNs.  The analysis
could be appreciated as a methodological synthesis of ideas from
Statistical Learning Theory \cite{Vapnik1998} and the Statistical
Field Theory \cite{mussardo2020statistical} in physics. We refer the
method as the {\bf statistical assembly method}. In this subsection,
we first introduce the statistical-field approach and the
statistical-learning approach, and discuss their limitations, and then
we introduce the statistical assembly method.

The statistical-mechanics approach to study DNNs have a long history
\cite{Seung1992,Watkin1993,Haussler1996,Carleo2019}, however, the
application of this approach typically requires homogenization of data
and models, and thus the ensued analyses are away from practical
settings. More specifically, in physical systems, a {\it field} is a
coarse-grained characterization of a collection of particles, and it
is a good formal model of the particles' collective behaviors because
the disorganized interaction among particles, unraveled through a
timescale orders of magnitude larger than the thermodynamic timescale,
results in statistically stable, homogeneous behaviors of this
collective. A typical example would be the mean-field models, which we
refer to \supps D D for a review. As a result,
statistics of the collective makes a coarse-graining for analysis,
which is also known as {\it effective theory} in physics. And more
particularly, the behaviors of particles could be characterized as
Gaussian fields/distributions, and thus the many-body interaction of
particles are high-dimensional Gaussian integrals. Therefore, to apply
the statistical field method to neural networks, the data and
interaction among neurons need to be homogenized
\cite{Seung1992,Watkin1993,Haussler1996,Carleo2019,Goldt2020,Goldt2021},
and the setting is away from practical setting \cite{Carleo2019}; for
example, the input data are assumed to uniformly sampled from a
hypersphere---more related works could be found in the related-works
discussion in \supps A D 4, where the works that do
finite-width correction to infinite-width assumption typically make
such homogeneous assumptions over data.

Meanwhile, the statistical learning theory is a revolution in
statistics that does not requires restrictive assumptions, such as
those in analyses from the approach of statistical mechanics, however,
the theory was developed to analyze relatively simple models, and the
analyses do not generalize straightforwardly to complex models in high
dimensional setting like DNNs---currently, complex models on high
dimensional data like DNNs lie in a ``no man's land'' between
efficient linear methods on high-dimensional data with strong
regularities in the sense of concentration-of-measure phenomena, and
low-dimensional data with efficient complex nonlinear methods
\cite{Gorban2018}. More specifically, the utilities of the
statistical-learning analysis lie in characterizing worst-case
behaviors that are close to the practical behaviors in the sense that
the former could qualitatively characterize the latter---and in many
cases, the worst case behaviors could prescribe quantities that
control generalization; this resembles control parameters in
statistical physics. Concretely, a primary goal of learning theory is
to characterize the generalization of a hypothesis/classifier through
an upper bound. The upper bounds obtained are worst case behaviors of
samples. In simple models, the behaviors characterized by the bounds
are close to practical behaviors, in the sense that, for example, the
bounds can qualitatively characterize the generalization ability of
the hypothesis by identifying quantities (e.g., margins of support
vector machine (SVM), or margins \cite{Bartlett2017}, distance from
initialization \cite{Neyshabur2018}, singular values \cite{Jia2019} of
DNNs) that qualitatively characterizes the generalization
ability. More concretely, with the margins of SVM as an example,
though each training example has a margin of its own, the bounds are
characterized by the smallest margin among all training
examples. Consequently, the generalization errors are over-estimated,
and the utility of the bound is to identify the margin as a
qualitative characterization of generalization, and in practice, the
margins of all examples would be intentionally maximized to achieve
better generalization. However, as probability bounds, their values
are typically much larger than $1$---the looseness of generalization
bounds in the context of DNNs has been discussed
\cite{Zhang2016b,Bartlett2017}. And to reach descriptive or
prescriptive bounds, extra care is needed to identify worst-case
behaviors that are close to practical behaviors. Otherwise, the
intuition obtained from simple models could be misleading: for
example, the bias-variance trade-off is based on analysis of simple
models, and for complex models like DNNs, the behaviors of
generalization are not exactly the same with the broad-stroke
bias-variance trade-off, and manifests as the double descent
phenomenon \cite{Belkin2019,Advani2020}.

In this work, we synthesize the two approaches through symmetries: we
do not make restrictive assumptions on data distributions as works
under the statistical-mechanics approach do, which is achieved by
analyzing worst-case behaviors similar to works in statistical
learning theory, and that we do calculate many-body interaction in the
system, but at the granularity of assemblies that characterize
hierarchical many-body interactions. This approach could be
appreciated biologically: given the unknown in an environment, a
strategy for an organism to prevent coincidental survival risk is to
hoard a reservoir of backup plans to stay far away from vulnerable
situations. More specifically, \citet{Carleo2019} speculate that the
statistical-mechanics approach and statistical-learning approach are
complementary once we understand the key conditions under which
practical cases are close to worst cases. In DNNs, the basic units are
not homogeneous particles, but heterogeneous basis circuits that each
has its particular behaviors. The coarse-grained behaviors of the
basis circuits are formalized as assemblies that also each has its own
particular behaviors. And it seems that it is only analytically
amenable to analyze the extreme case behaviors of all assemblies---this
is a trade-off made between verisimilitude and analytical
tractability. As a result of this trade-off, the quantitative
characterized achieved would be significantly lower or higher
(depending on whether a lower or an upper bound is taken) than the
actual behaviors of the system.  Meanwhile, circuit symmetry
formalizes the repetitive/invariant behaviors across assemblies. Such
invariance makes the extreme-case behaviors close to the practical
behaviors.  Consequently, although the quantitative characterization
in this work does {\it not exactly} characterize the behaviors of
DNNs, it does so {\it qualitatively}, as in the approach of
statistical learning theory. Meanwhile, the extreme-case
characterization is calculated from extreme-case behaviors of
many-assembly interaction up to arbitrary order, and thus it is also a
statistical characterization resembling that of the statistical field
theory.  For conceptual clarity, we refer this method as statistical
assembly methods, to clarify both the difference and the similarity
with the statistical learning theory and statistical field theory.

The rest of this work is the application of the method to study the
coarse-grained effect of circuit symmetries that leads to a phase of
DNNs: the correlations among assemblies of all orders are calculated,
and when a DNN is hierarchically large, a concentration-of-measure
phenomenon manifests where the eigenspectrum of a DNN's Hessian could
be characterized by a self-consistent matrix equation, and that
qualitatively characterizes practical behaviors.

\subsection{Order from fluctuations, or order from adaptive symmetry}
\label{sec:order-from-fluct}

The circuit symmetry introduced in \cref{sec:adapt-symm-feedb} is a
heterogeneous symmetry, which implies, as we have explained in
\cref{sec:adapt-symm-feedb}, that during the self-organizing of DNNs,
intact and broken symmetries could coexist: broken symmetries encode
information about the environment/dataset, and manifest as complex
functional structure; the intact symmetries
maintain the ability to further reduce information perturbations
(training errors). Therefore, the self-organizing process is a process
where circuit symmetries are continually being broken. Further recall
that circuit symmetries characterize the symmetric
distribution/perturbations of (output of) basis circuits around
zero. Therefore, the preceding self-organization (risk minimization)
is a salient ``order from fluctuations'' phenomenon that characterizes
self-organization, and the order emerges from disorder in circuits by
selective (positive, or negative) feedback signals that breaks the
circuit symmetries. However, to further study such a symmetry-breaking
process, we need to clarify the concept of {\it order} in DNNs next,
and a more rigorous presentation and further details are given in
\supps B F.

\subsubsection{Order in physics and self-organization}
\label{sec:order-physics-self}

To begin with, we introduce the concept of order in physics.  In
\cref{sec:adapt-symm-feedb}, we have explained that symmetries
identify observables that are coarse-grained effect of symmetries.
Intuitively, for a system consisting of microscopic units to manifest
behaviors that could be consistently observed at a macroscopic scale,
a synchronization/cooperation of a large number of the microscopic
units needs to somehow occur. And the synchronization results in {\it
self-similarity} between the microscopic units and the
macroscopic/system-wide behaviors. And in physics, such a macroscopic
behavior is referred as the {\bf order} of a system. The
self-organization of a physical system is the process where the
constituent units synchronize to transit from one macroscopic behavior
to another. To give an example in daily life, snowflakes also
self-organize: when temperature falls, the rotational symmetry is
broken into a six-fold symmetry, and macroscopically, and we observe
the flower-like shape of the snowflakes. The shape is the
(macroscopic) order of water molecule system emerges from disorder at
high temperature, and the temperature is the control parameter of the
process.

\begin{figure*}[t]
  \centering
  \subfloat[\label{fig:op_stats}]{%
  \includegraphics[width=0.49\linewidth]{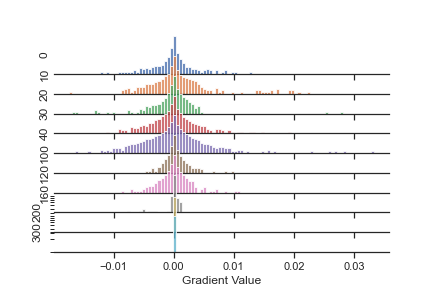}%
  }
  \subfloat[\label{fig:op_curve}]{%
  \includegraphics[width=0.49\linewidth]{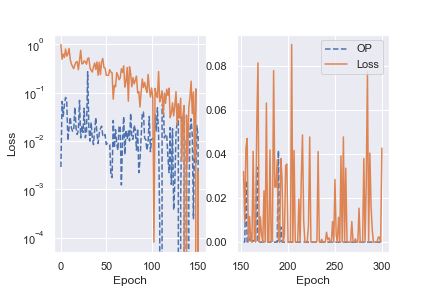}%
  }
  \caption{Adaptive symmetry of, or coarse-grained effect of circuit
symmetries on gradients, and the dynamics of the plasticity order
parameter. Gradients are perturbation of neuron assemblies of order
one. And they are experimentally found to be self-similar to the
circuit symmetries of perturbation of basis circuits of order one
throughout training (shown in fig. a), implying that the
coarse-graining of the mixture of intact and broken circuit symmetries
manifests as adaptive symmetries at the assembly level. The further
coarse-graining of gradients by adding up the square of gradients
leads to nonzero squared norm of the gradient vector, and
characterizes that whether the coarse-grained effect of circuit
symmetries is capable of breaking to decrease the errors (measured by
empirical risk) of the prediction of the coarse-grained variable
(i.e., $H_L$ in \cref{fig:dnn_graph}) computed by the network (shown
in fig. b). Thus, according to the epistemology of physics, the
squared norm of gradient vector is referred as the plasticity order
parameter to quantitatively characterize the order of DNNs. {\bf (a)}
Adaptive symmetry of gradients throughout DNN training. The figure
plots the distributions of gradients throughout training, where the
numbers on the left of the plot are the epochs where the gradients are
computed. {\bf (b)} The transition of plasticity order parameter and
empirical risk from nonzero to zero during DNN training. The curve
denoted by {\tt OP} is the step-size rescaled plasticity order
parameter throughout training. The curve denoted by {\tt Loss} is the
training loss/risk throughout training. The plot is divided into two
subplots because we plot the first 150 epochs in logarithmic scale to
emphasize that the scaled order parameter decreases roughly by the
same rate with the loss, and is approximately an order of magnitude
smaller than the loss; and in the second subplot, we need to revert
back to the normal scale because the loss is too small and as such,
the statistical fluctuations are amplified in logarithmic scale, and
obfuscate the information that both the order parameter and the loss
are close to and oscillate to zero.}
  \label{fig:order_parameter_exp}
\end{figure*}

The order described previously is a macroscopic behavior of a physical
system at {\it equilibrium}. And when a system is referred as
self-organizing, it tends to describe a {\it non-equilibrium}
system. And the order of a self-organizing system typically refers to
a stable dynamic behavior, for example, a dissipative structure such
as the Rayleigh-Bénard convection, which emerges from stable
fluctuations caused by energy input/gradients. Such dynamics is
summarized as ``order from fluctuations'' by \citet{Prigogine1978},
``order from noise'' by \citet{VonFoerster2003}, or ``order from
chaos'' by \citet{kauffman1993origins}. The circuit-symmetries
breaking presented in \cref{sec:adapt-symm-feedb} is exactly such a
process: each basis circuit fluctuates (is of probability measure
distributed) symmetrically against $y$-axis to break its symmetry to
decrease the error of novel examples.

\subsubsection{Order of DNNs}
\label{sec:order-dnns}

We characterize the order of DNNs, by studying the coarse-grained
effect of circuit symmetries introduced in
\cref{sec:adapt-symm-feedb}. We shall present a macroscopic adaptive
symmetry that characterizes the order of DNNs: a self-similarity
between the macroscopic (i.e., neuron assemblies) and the microscopic
(i.e., basis circuits composed by neurons).

First, we present an {\it order parameter} of DNNs. In statistical
physics, the order of a system is formally and quantitatively
characterized by order parameters that coarse-grain the symmetric
behaviors of microscopic units, e.g., the magnetization order
parameter of magnets. The order parameter of a system is typically
identified by examining the {\it elementary excitation}
\cite[Chp~9.3]{entropy} of symmetries in the system in response to
perturbations: the magnetization order parameter of a ferromagnet is
the first-order excitation of magnetic spins in response to an
external magnetic field, or colloquially, putting a magnet close to
magnetic field, the collective behaviors of the units in the magnet
would induce a repulsive or attractive force, depending on the
polarity---a review of these concepts in physics is given in
\supps D C. The external perturbations of a DNN
are simply novel examples of nonzero errors, and we examine the
elementary excitation of DNNs in response to novel examples and
formally define the first order excitation (i.e., the square of
gradient norm) as the order parameter of DNNs,
\begin{equation}
  \label{eq:circuit_form_gradient_norm-pre}
  ||\nabla_{\theta}R||_2^{2} = \sum_{l=1}^{L}\sum_{i_{l-1}=1,i_{l}=1}^{n_{l-1}, n_{l}}
    \sum_{\bv{i}, \bv{i}' \in I^{\setminus \{l-1, l\}}_{i_{l-1}i_{l}}}
     \Omega^{\overl{l}_{\setminus \{l\}}}_{\bv{i}} \Omega^{\overl{l}_{\setminus \{l\}}}_{\bv{i}'},
\end{equation}
where $\overl{l}$ denotes $(1\ldots L)$, $\Omega^{\overl{l}_{\setminus \{l\}}}_{\bv{i}}$
denotes
\begin{equation}
  \label{eq:first_derivative_basis_circuit-pre}
  X_{i_0}\Psi^{\wedge 1\sim l-1\wedge}_{\bv{i}_{:l}}H^{l-1}_{i_{l-1}}H^{l}_{i_l}\Psi^{\wedge l+1\sim L\wedge }_{\bv{i}_{l:}}\ca{L}'(\cdot),
\end{equation}
which is the first derivative of basis circuit given at
\cref{eq:basis-circuit-pre} that intakes a datum $X_{i_0}$ at the
bottom layer, and a derivative $\ca{L}(\cdot)$ at the top layer; and $I^{\setminus
\{l-1, l\}}_{i_{l-1}i_{l}}$ denotes
\begin{displaymath}
  \overrightarrow{\otimes}_{\{p \in \overl{l} | p < l-1\} }[n_{p}] \otimes \{i_{l-1}\} \otimes \{i_{l}\}
  \overrightarrow{\otimes}_{\{q \in \overl{l} | q > l\} }[n_{q}];
\end{displaymath}
    that is, $[n_{l-1}], [n_{l}]$ are substituted by single element
sets $\{i_{l-1}\}, \{i_{l}\}$, respectively, in the consecutive
Cartesian products that previously composes the index set $I$ of basis
circuits in \cref{eq:basis-circuit-pre}. We refer $\Omega^{\overl{l}_{\setminus
\{l\}}}_{\bv{i}}$ as \textbf{first order perturbations of basis
circuits induced by weights}---an illustration is given in
\cref{fig:dnn_graph}---and \cref{eq:first_derivative_basis_circuit-pre}
the {\bf plasticity order parameter}, which characterizes whether a
DNN is able to continue to absorb information perturbations, or in
other words, to decrease the errors of novel examples.

Remarkably, under the circuit formalism, the plasticity order
parameter is symbolically equivalent with the spin glass order
parameter.  Recall that the spin order parameter of spin glasses is
\begin{equation}
  \label{eq:spin_glass_order_parameter_pre-pre}
  \propto \bb{E}_{\rho(\bv{\sigma})}[\sum_{ij}\sigma_i\sigma_j],
\end{equation}
where $\sigma_{i}, i \in [n]$ denotes spins, and $\rho$ denotes the law of the
spins $\bv{\sigma}$---it could be derived by taking the first order
derivative of free energy (i.e., the elementary excitation), and the
derivation is given in \supps D C for interested
readers. \Cref{eq:circuit_form_gradient_norm-pre} resembles
symbolically to \cref{eq:spin_glass_order_parameter_pre-pre} if we
merge the three summation symbols into one big symbol. The resemblance
is also semantically informative to understand the coarse-grained
symmetries of DNNs.  Recall that in \cref{sec:dnn-self-organ}, we
interpret the risk minimization of DNNs as a
variational-free-energy-minimization process where the uncertainty of
the coarse-grained variable that detects macroscopic patterns is
minimized by a feedback-control loop between the coarse-grained
variable and the hierarchical circuits computing the variable. The
plasticity order parameter is the quantified uncertainty reduced in
one step of the risk minimization, and is a coarse-grained variable
that coarse-grains the change in the neuron circuits that induces the
uncertainty reduction.  A spin glass self-organizes to align spins at
a macroscopic scale; thus order manifests when the spins align. A DNN
self-organizes to reduce uncertainty in pattern recognition; thus
order manifests when the basis circuits pose to reduce
uncertainty. The coarse-graining of spins computes the degree of
alignment/order that induces magnetic force/potential in response to
external magnetic field, whereas the coarse-graining of the output of
circuits computes the potential reduction of uncertainty in response
to information perturbations. Both spins and basis circuits make
binary choices that increase or decrease the order parameter, though
the binary choices of circuits result from the hierarchical coupling
of many neurons.

Notice that the differentiation only removes one weight from each
basis circuit, and thus if the weight symmetry still holds for the
rest of the weights, circuit symmetry still holds.  A self-similarity
(of adaptive symmetry) between gradients and first-order perturbation
of basis circuits is experimentally observed in \cref{fig:op_stats}:
each entry of the gradient is a coarse-grained observable of
perturbation of basis circuits (i.e., neuron assemblies),
\begin{displaymath}
    \sum_{\bv{i} \in I^{\setminus \{l-1, l\}}_{i_{l-1}i_{l}}}
    \Omega^{\overl{l}_{\setminus \{l\}}}_{\bv{i}},
    l=1\ldots L, i_{l-1}=1\ldots n_{l-1}, i_{l} = 1\ldots n_{l};
\end{displaymath}
they are observed to be symmetrically distributed against $y$-axis,
and gradually converge to zero during the training of a
DNN. Correspondingly, as could be seen in \cref{fig:op_curve}, the
plasticity order parameter gradually transits from nonzero to zero,
and reaches zero when the empirical risk is zero---recall that in
physics, the order parameters transits from zero to nonzero (or, vice
versa) during phase transitions.

\subsection{Plasticity phase, extended symmetry breaking, and benign pathways on the risk landscape}
\label{sec:plast-phase-benign}

In \cref{sec:order-from-fluct}, we explained that the order of a
system is the macroscopic behaviors of the system that are the
coarse-grained effect of stable symmetries in the system, and are
quantitatively characterized by order parameters. And thus different
{\it stable} symmetries would induce distinctive macroscopic
behaviors, and such distinctive behaviors are referred as {\it phases}
of the system. In this section, we shall characterize a phase of DNNs,
which we refer as the {\it plasticity phase}, where a coarse-grained
adaptive symmetry exists that is self-similar to the circuit symmetry
of basis circuits. Meanwhile, the phase is both a phase of a DNN,
where the DNN could continually decreases risk, and an {\it extended
symmetry-breaking} process, where the symmetries are continually being
broken during the self-organizing process.  As a result, the {\it
benign pathways} (that lead to zero risk) exist, and thus suggests an
explanation of the marvelous optimization power of DNNs in
practice. We introduce the phase in the following, and a more rigorous
presentation and further details are given in
\supps B G.

\subsubsection{Phase space in physics}
\label{sec:phase-space-physics}

To begin with, we clarify the concept of {\it phase space} in physics.
Recall that in \cref{sec:adapt-symm-feedb}, we explained that symmetry
groups formalize invariants of physical systems that constituent the
fundamental concepts to understand these systems. A formalism that
systematically organizes the concepts is the phase space: in
statistical physics, a phase space is a mathematical space whose
coordinates are control parameters, and order parameters are functions
defined on the space that uniquely determine the system's behavior of
interest \cite{Longo2013}---both the control and order parameters are
observables identified by symmetries.  Further recall that in
\cref{sec:order-from-fluct}, we explained that order parameters
characterize the system's excitation in response to
perturbations. When the energy perturbation is sufficiently large or
the system is at critical regimens, the symmetries would be broken,
and the system would undertake a phase transition from a phase with
one set of stable symmetries to another. The phase transitions---that
is, symmetry breaking---are singular points of the function where the
order parameter singularly changes from one value to another. For
example, the phase space/diagram of spin glasses is a Euclidean space
(cf. figure 2.1 in \citet{nishimori01}) whose coordinates are control
parameters (e.g., temperature); and order parameters (e.g., spin glass
order parameter) are functions defined on the space that uniquely
determine the system's behavior of interest (i.e., paramagnetic,
ferromagnetic, or glassy phase). Stable rotation and translation
symmetry of magnetic spins in a spin glass is referred as the
paramagnetic phase, and is quantitatively characterized by zero
magnetization order parameter. The phase transitions from the
paramagnetic phase to the ferromagnetic phase---that is, rotation
symmetry breaking---are singular points of the (order parameter)
function where the magnetization order parameter singularly changes
from zero to nonzero.  More detailed review of the concepts is given
in \supps D C.

\subsubsection{Hypothesis on a phase of DNNs with stable circuit symmetries}
\label{sec:hypoth-phase-dnns}

Following the epistemology of physics, to understand the symmetry
breaking of DNNs, we investigate the phases of DNNs. We hypothesize
that the consistent adaptive symmetry of gradients experimentally
observed in \cref{sec:order-dnns} is a phase of DNNs, i.e., a stable
macroscopic behaviors of DNNs. More specifically, recall that in
\cref{sec:adapt-symm-feedb}, we introduce that because circuit
symmetry is a heterogeneous symmetry, the circuits in a DNN could be
of broken symmetry in only a subset of all circuits, and the system is
in a state where intact and broken circuit symmetries coexists. And in
\cref{sec:order-from-fluct}, we introduce the experiments that observe
a macroscopic adaptive symmetry self-similar to adaptive symmetry of
basis circuit (i.e., circuit symmetry): the gradients as
coarse-grained observables of perturbation of basis circuits are
symmetrically distributed against y-axis throughout the training of a
DNN. Therefore, we might hypothesize that throughout the training, the
DNN is in states where circuit symmetries stably exist (along with
broken circuit symmetries) to manifest adaptive symmetry at the level
of assemblies (e.g., symmetry of the gradient distribution). And the
adaptive symmetry of assemblies mentioned in
\cref{sec:order-from-fluct} is potentially a phase of DNNs that is
characterized by the nonzero of the plasticity order parameter.

\begin{figure}
  \centering
  \includegraphics[width=0.8\linewidth]{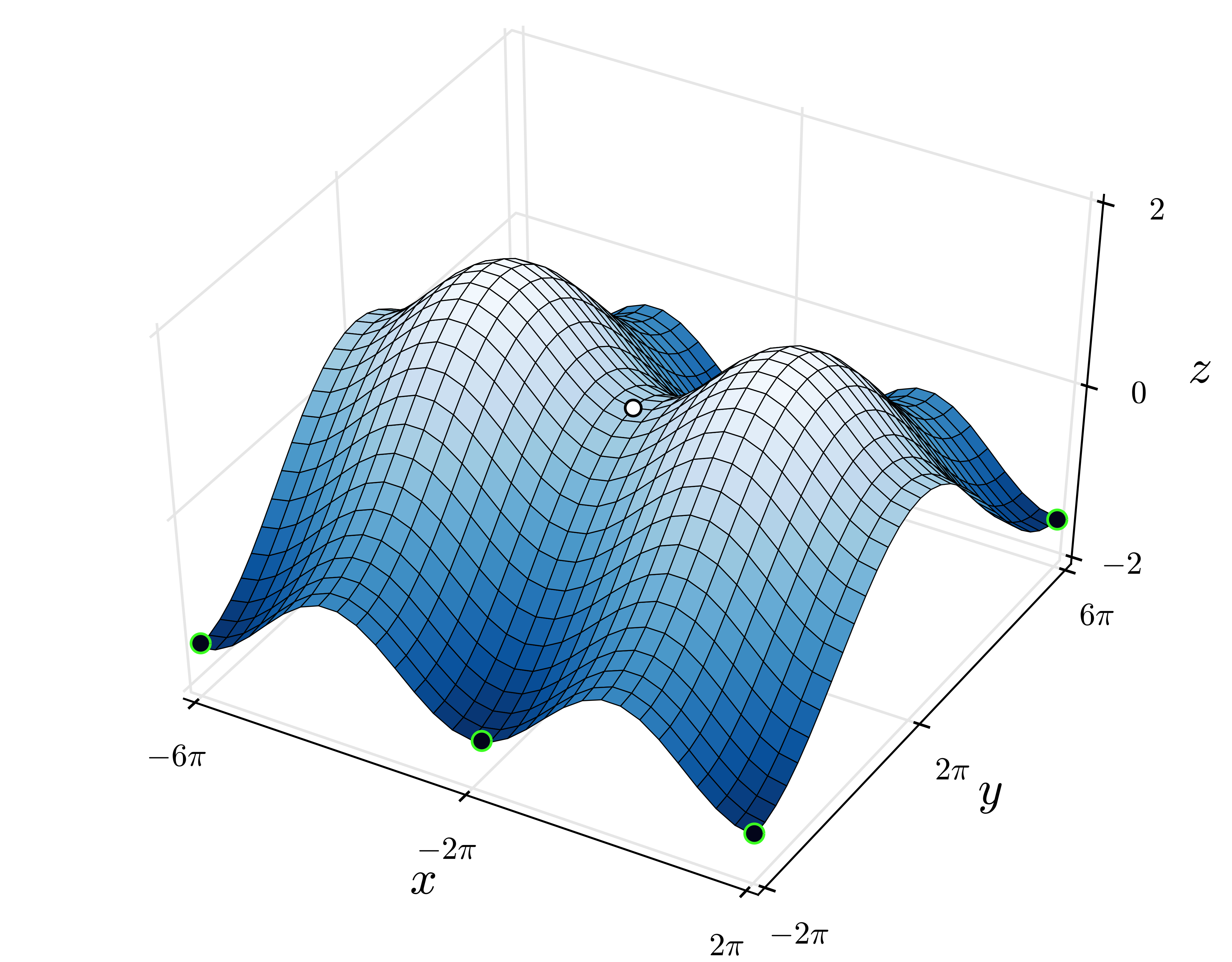}
  \caption{Illustration of saddle points and a functional landscape where local minima are global
    minima; the white
    dot in the middle, and black dots at the corners are a saddle point, and
    local/global minima, respectively. }
  \label{fig:landscape}
\end{figure}

\subsubsection{Symmetry-stability analysis through Hessian}
\label{sec:study-stabl-circ}

To investigate the preceding hypothesis, we analyze the stability of
the circuit symmetries. The stability analysis is performed by
analyzing the Hessian of DNNs, as introduced in the following.

In physics, as order parameters are associated with the first order
symmetric excitation of units in response to excitation, the stability
of order is characterized by the second order effect of such
excitation (e.g., magnetic susceptibility).  Thus, the stability of
symmetries is characterized through Hessian of the free energy---a
review of Hessian analysis of phases in physics could be found in
\supps D E.  Hessian of a potential function
informs us of the local geometry of the risk landscape: when the
gradient at a state is zero vector, negative eigenvalues of the
Hessian inform that at the current state, descent directions in the
parameter space exist to further reduce the potential; otherwise, the
current state is a local minimum. A illustration of local geometry of
risk landscape is given in \cref{fig:landscape}. Thus, to study the
stability of the coarse-grained effect of circuit symmetry, we need to
study the coarse-grained effect of circuit symmetries on the Hessian
of the variational free energy.

To further clarify, to appreciate the study of stability of symmetries
under the context of DNNs, note that the nonzero of plasticity order
parameter observed in \cref{sec:order-from-fluct} results from
symmetric statistical fluctuations of basis circuits that are only
weakly dependent on the other basis circuits, however, these are
experimental results obtained by sampling the order
parameters. Theoretically, despite the speculative weak dependence, it
could still be zero probabilistically. More concretely, geometrically,
recall that the self-organizing process is a gradient descent process
on the risk (variational free energy) landscape---recall that in
\cref{sec:dnn-self-organ}, we have explained that risk is also
variational free energy. Thus, the local curvature of the risk
landscape could be irregular, and probabilistically, the order
parameter could still be zero and the system could still stick trapped
at nonzero local minima.

\subsubsection{Coarse-grained effect of circuit symmetries on Hessian entries}
\label{sec:coarse-grain-effect-3}

\begin{figure*}[t]
  \centering
  \subfloat[\label{fig:hessian_entry_distributions}]{%
    \includegraphics[width=0.4\linewidth]{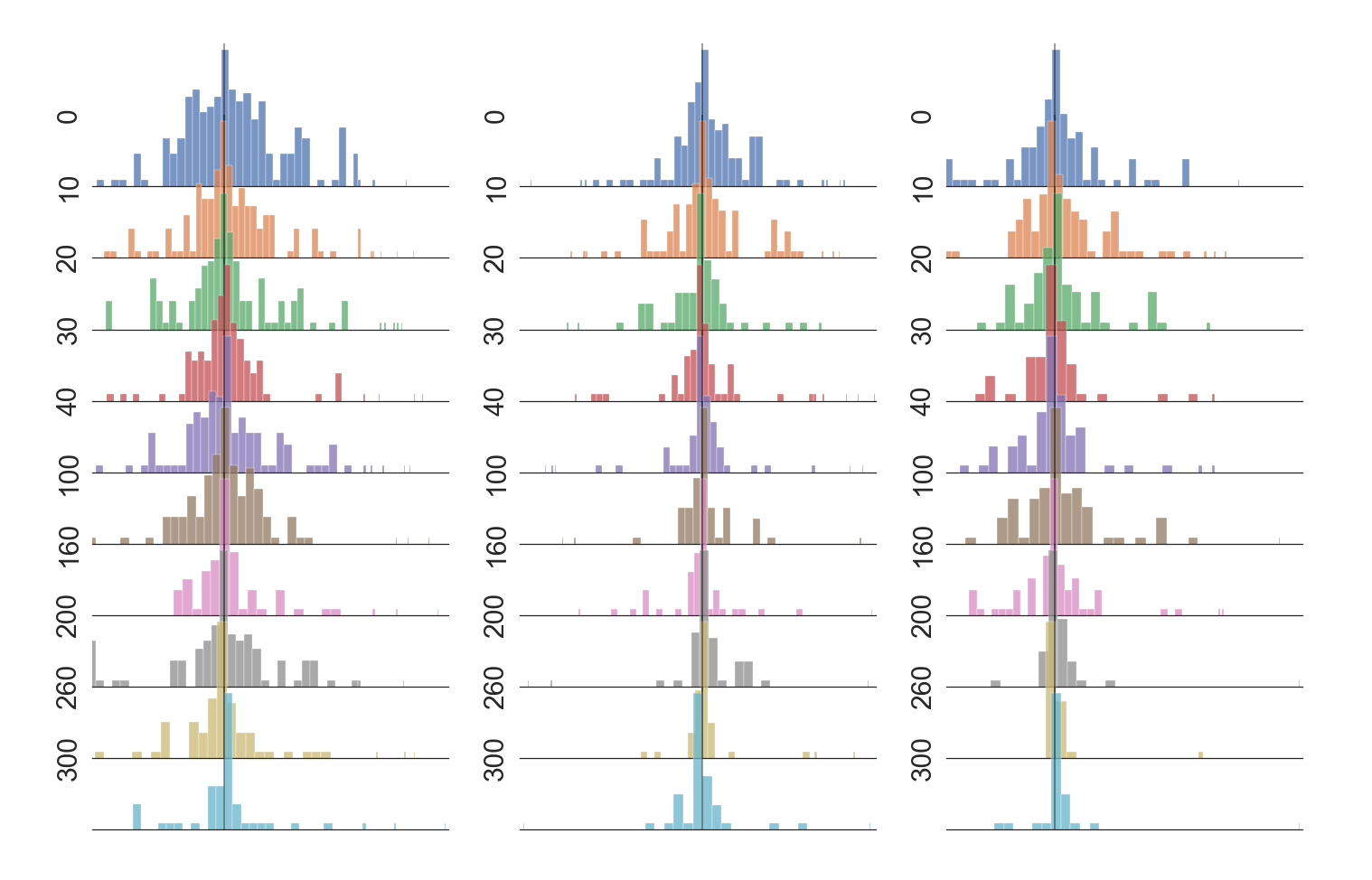}
    }
  \subfloat[\label{fig:cm_start}]{%
  \includegraphics[width=0.3\textwidth]{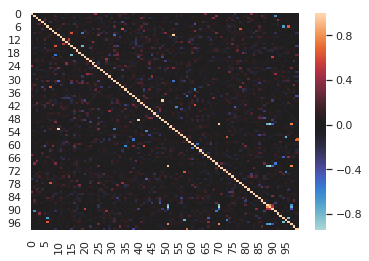}%
  }
  \subfloat[\label{fig:cm_end}]{%
  \includegraphics[width=0.3\textwidth]{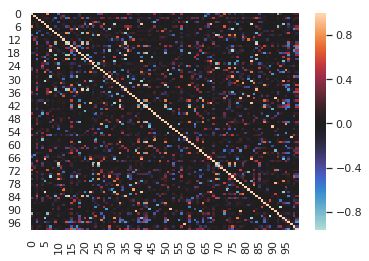}%
  }
  \caption{Adaptive symmetry of Hessian entries throughout training,
and sparse correlation among Hessian entries at the beginning and the
ending of DNN training.  {\bf (a)} Hessian entries are perturbation of
neuron assemblies of order two. They are also experimentally found to
be approximately self-similar as the gradients do. More specifically,
the figure shows the probability density distributions of three
randomly sampled Hessian entries, from left to right, throughout
training respectively. The numbers on the left are the epoch numbers
during training where the entry is sampled. Though we only show three
entries for demonstration, we have sampled many more entries and the
behaviors are similar. Both the x-axis and the y-axis are in the
logarithmic scale.  The Hessian entries are normalized by their
standard deviations, respectively; that is, the normalization is
entry-wise: each Hessian entry is divided/normalized by its standard
deviation to compare it with the fluctuations of this entry.  The
probability distribution of Hessian entries are computed with
bootstrap methods, of which the details is collected in
\supps I A. The basic idea is to repeatedly sample
the quantities of interest, and so long as the quantities under
sampling are of limited variations, the sample would serve as an
estimation of the quantities. In retrospect, circuit symmetries have
enabled us to observe stable behaviors by such a sampling method,
which is similar to the phenomenon that translation symmetry enables
stable experimental observations for repeated experiments performed at
different geo-locations. {\bf (b) (c)} are the correlation coefficient
(CC) matrices of Hessian entries at the beginning and the end of
training, respectively. The diagonal line is the variance, which is 1
as the result of normalization by standard deviation.  We uniformly
randomly subsample a $100 \times 100$ CC submatrix from a large CC
matrix---the small size $100$ is chosen to let the readers be able to
distinguish the pixels in the images. The CC matrix is obtained from
the bootstrap sample of Hessian entries noted in
\cref{fig:hessian_entry_distributions}. }
 \label{fig:exp_on_conditions}
\end{figure*}

The coarse-grained effect of (partial broken) circuit symmetries
results in the statistically stable assembly-level adaptive symmetry
(cf. the symmetric distribution of gradients in
\cref{sec:order-from-fluct}, which are also neurons assemblies), which
we discuss next.

First, as in the case of the order parameter where the gradients are
neuron assemblies that coarse-grains over the perturbation of basis
circuits of order one, the Hessian are neuron assemblies that
coarse-grains over the perturbation of basis circuits of order two.
More specifically, the second-order perturbation of risk in the
circuit form is given as
\begin{equation}
  \label{eq:dHd_circuit-pre}
  \begin{aligned}[t]
       \delta \bv{\theta}^{T}\bv{H} \delta \bv{\theta} = & \sum_{l=1}^{L}\sum_{l'=1}^{L}\sum_{i_{l-1}=1}^{n_{l-1}}\sum_{i_{l}=1}^{n_{l}}\sum_{i_{l'-1}=1}^{n_{l'-1}}\sum_{i_{l'}=1}^{n_{l'}} \\
                                     & \sum_{\bv{i} \in I^{\setminus \{l-1,l,l'-1,l'\}}_{i_{l-1}i_{l}i_{l'-1}i_{l'}}}\bb{E}[\Omega^{\overl{l}_{\setminus \{l, l'\}}}_{\bv{i}}] \delta w^{l}_{i_{l-1}i_l}\delta w^{l'}_{i_{l'-1}i_l'},
  \end{aligned}
\end{equation}
where $\bv{H}$ denotes the Hessian matrix of risk, and $\delta\bv{\theta}^{T}$
is vectorized weight perturbations,
\begin{equation}
  \label{eq:2-order-perb-bs}
  \Omega^{\overl{l}_{\setminus \{l, l'\}}}_{\bv{i}} := \Psi^{1\sim l-1}_{\bv{i}_{:l}}\Psi^{l+1\sim l'-1}_{\bv{i}_{l:l'}}\Psi^{l'+1\sim L}_{\bv{i}_{l':}},
\end{equation}
and $I^{\setminus \{l-1,l,l'-1,l'\}}_{i_{l-1}i_{l}i_{l'-1}i_{l'}}$ denotes
    \begin{displaymath}
      \begin{aligned}[t]
      &\overrightarrow{\otimes}_{\{p \in \overl{l} | p < l-1\} }[n_{p}] \otimes \{i_{l-1}\} \otimes  \{i_{l}\}\overrightarrow{\otimes}_{\{p\in \overl{l} | l < p < l'-1\} }[n_{p}]\\
      &\otimes \{i_{l'-1}\} \otimes \{i_{l'}\} \overrightarrow{\otimes}_{\{p \in \overl{l} | p > l\} }[n_{p}].
      \end{aligned}
    \end{displaymath}
Although the equations look complicated, \cref{eq:dHd_circuit-pre} is
simply the coarse-graining of the second-order perturbation of basic
circuit $\Psi^{\overl{l}}_{\bv{i}}$ induced by the perturbations of
circuit weights at layer $l, l'$, i.e., $\delta w^{l}_{i_{l-1}i_l}\delta
w^{l'}_{i_{l'-1}i_l'}$---it is also explained with an illustration
previously in \cref{fig:dnn_graph}.

Second, as a result of heterogeneous circuit symmetries, even if a
substantial subset of the circuits are of broken circuit symmetries,
the coarse-grained behaviors of circuits could still be close to in
the case where all circuit symmetries are intact. To illustrate the
phenomenon, we present the upper bounds on the moments of assemblies
as follows.  Let $\{\delta^{k}\Psi^{1\sim L}_{\bv{i}}\}_{\bv{i} \in \ca{B}_k}$ be
perturbation of basis circuits, and $\ca{B}_0, \ca{B}_1, \ca{B}_2$
denotes $I, I^{\setminus \{l-1, l\}}_{i_{l-1}i_{l}}, I^{\setminus
\{l-1,l,l'-1,l'\}}_{i_{l-1}i_{l}i_{l'-1}i_{l'}}$, respectively.
Suppose that $\forall n_l = n \in \bb{N}^{+}$, and there exists $\mu > 0, \mu \in
\bb{R}$, such that $\sqrt{n}^{1-\mu}$ weights are of broken weight
symmetry at each layer, then we have
  \begin{equation}
    \label{eq:T_coarse-grained-broken-pre}
    \begin{aligned}[t]
    &\bb{E}[\sum_{\bv{i} \in \ca{B}_k}\delta^{k}\Psi^{1\sim L}_{\bv{i}}] \leq O\left( (\frac{c^2}{2})^{L} + \frac{1}{\sqrt{n}^{L\mu}}c^{L} \right),\\
    &\bb{E}[(\sum_{\bv{i} \in \ca{B}_k}\delta^{k}\Psi^{1\sim L}_{\bv{i}})^2] \leq O\left( c^{2L} + (\frac{c^2}{2})^{2L} + \frac{1}{\sqrt{n}^{2L\mu}}c^{2L} \right),
    \end{aligned}
  \end{equation}
where $\delta^{k}\Psi^{1\sim L}_{\bv{i}}$ is a shorthand notation for $\Psi^{1\sim
L}_{\bv{i}}, k=0$, $\Omega^{\overl{l}_{\setminus \{l\}}}_{\bv{i}}, k=1$,
$\Omega^{\overl{l}_{\setminus \{l, l'\}}}_{\bv{i}}, k=2;$ the derivation is in the
\supps B E 6---note that the assembly ($k=2$) here
computes a Hessian entry while \cref{eq:dHd_circuit-pre} further
coarse-grains all Hessian entries. Notice that for a hierarchically
large DNN, $\sqrt{n}^{L}$ and $2^{L}$ are both very large value, and
thus the means are close to zero, and the squared norm of assemblies
are close to $c$ (which is close to $1$ through initialization---this is
a carefully maintained edge-of-chaos condition, and interested readers
could find more discussion in \supps C B).  The
decay of mean is a special case of higher order moments where weights
of odd power exists: when calculating the odd moments of a basis
circuit $H^{L}_{i_{L}}W^{l}_{i_{l-1} i_l}\Psi^{1\sim L-1}_{i_{1}\ldots i_{L-1}}$,
the integration against $W^{L}_{i_{l-1} i_l}$ over $[-\infty, 0]$ and $[0,
+\infty]$ would almost cancel out, and only leaves a residual because
weights and neuronal gates are sparsely and weakly correlated; further
notice that odd powered weights at each layer would induce a decay,
and thus an exponential decay against depth is induced. Further notice
that for higher moments, the majority of cross-moments among basis
circuits (e.g., $\Psi^{1\sim L}_{\bv{i}}\Psi^{1\sim L}_{\bv{j}}\Psi^{1\sim L}_{\bv{k}}$)
have odd power weights. Therefore, we might speculate a close-to-zero
mean and sparse correlation among Hessian entries.

\begin{figure*}[t]
  \centering
  \subfloat[\label{fig:hessian_mean_distributions}]{%
  \includegraphics[width=0.3\linewidth]{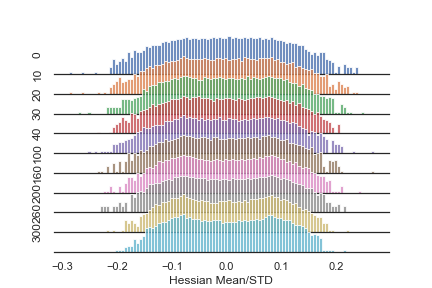}%
  }
  \subfloat[\label{fig:zero_fraction_throughout_training}]{%
  \includegraphics[width=0.3\linewidth]{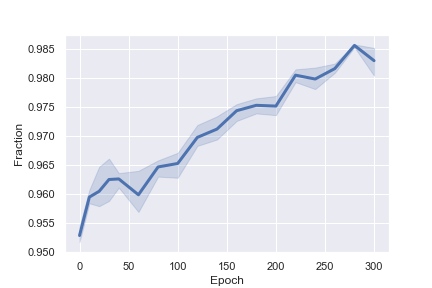}%
  }
  \subfloat[\label{fig:control_parameter_curve}]{
    \includegraphics[width=0.3\linewidth]{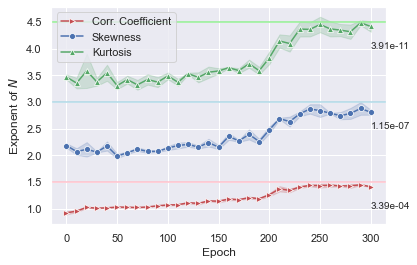}
  }
  \caption{Experimental support for the assembly-symmetry
assumption. {\bf (a)} The probability density distribution of means of
Hessian entries throughout training. The entries are also obtained by
the bootstrap method summarized in
\cref{fig:hessian_entry_distributions}, and the means are normalized
by standard deviations of Hessian means to weep out the influence of
magnitude throughout training and reveal the consistent symmetry. {\bf
(b)} The fraction of standard-deviation-normalized means that are less
than 0.15 throughout training. The transparent region is the standard
deviations of the fractions estimated through the bootstrap methods
summarized in the caption of
\cref{fig:hessian_mean_distributions}---because the fractions are
estimated through the bootstrap method, the standard deviations are
plotted to show that the estimated fractions are stable estimations.
{\bf (c)} Experimental support for the assembly-diversity and
assembly-diversity assumption.  The average order parameter, i.e.,
coupling set size, (over all Hessian entries) is estimated through the
average non-zero statistics (of all Hessian entreis), i.e.,
correlation coefficients, skewness, and excess kurtosis, in term of
the exponent of Hessian dimension $N$ during training, shown as curves
in the figure. {\it The non-zero correlation coefficients of each
entry is equivalent to its coupling set size:} the details on the
relationship between the number and coupling set sizes is given in
\supps B G 7. The transparent regions are standard deviations from the
bootstrap samples in the experiments. The light color straight lines
and the annotated numbers in the rightmost indicate the ratio between
the non-zero number and the overall number of the statistics (both
zero and non-zero). The numbers are estimated from bootstrap methods
described in \supps I B.}
  \label{fig:hessian_mean}
\end{figure*}

Indeed, we experimentally compute mean and correlations among Hessian
entries, and found that the Hessian entries fluctuate around zero and
are of sparse correlations among one another, which could be
appreciated as adaptive-symmetry and diversity of neuron assemblies,
as explained in the following.
\begin{enumerate}[leftmargin=0cm]
\item The adaptive symmetry of biotic systems has been introduce in
\cref{sec:adapt-symm-feedb}, and the adaptive symmetry of Hessian
entries manifest as the phenomenon that the Hessian entries fluctuate
around zero such that random fluctuations could be converted into
system-wide asymmetric behaviors in response to feedback signals---that
is, the system poses to adapt symmetrically.
\item And in biotic systems, diversity begets complexity
\citep{10.5555/1984802}: biotic systems evolve from low fitness
states---relatively speaking---through a series of adjacent states to high
fitness states on the frustrated fitness landscape by increasing
diversity of the system. And such diversity exists in neuron
assemblies as well. At initialization, DNNs are in a disordered state
where the weights among neurons are independently randomly sampled;
that is, a set of neurons with a diversity of possible
coupling. Accordingly, different signals would activate a different
set of neurons in the network randomly. During risk minimization by
gradient descent, the weights are adjusted by the gradient computed by
back-propagation until convergence. For a hierarchical large network,
the number of possible paths (i.e., basis circuits) among neurons grow
exponentially with depth, and thus at each weight update, the
information only flows within a subset of neurons that are small
compared with the overall number of possible
circuits. Correspondingly, the after the training, though for a given
signal, circuits are formed to classify it, for different signals,
neurons still are activated largely uncorrelatedly and thus {\it
seemingly randomly}. And thus the diversity of weights is a stable
characteristic, and induces a mixture of order and disorder through
training. This mixture manifests at the assembly level as functional
behaviors (e.g., the prediction of a DNN) of, and the sparse
correlation among assemblies: the seemingly random activation of
assemblies would result in low correlation among them. And thus the
Hessian entries are of sparse correlation among one another.
\end{enumerate}

Experimentally, from \cref{fig:hessian_entry_distributions}, we could
see that the Hessian entries are distributed approximately
symmetrically against $y$-axis---a self-similar adaptive symmetry
between Hessian entries (i.e., neuron assemblies of order two) and the
microscopic basis circuits, as that of the gradient (i.e., neuron
assemblies of order one), and thus fluctuate around zero throughout
training. From \cref{fig:cm_start} and \ref{fig:cm_end}, we could see
that the correlations among Hessian entries are rather sparse, and the
correlations increases after training, presumably because the circuit
symmetries are broken to encode information in the dataset, and thus
the statistical dependence among Hessian entries
increase. \Cref{fig:hessian_mean} shows that throughout the training,
the means concentrate on zero in the sense that most of the means do
not fluctuate further than $0.15$ standard deviation from zero, which
is clearly a concentration of measure on zero, more quantitatively,
from \cref{fig:zero_fraction_throughout_training} we can see that
throughout training, the fraction of the means that are less than
$0.15$ is at least $95\%$, and as the training progresses, the
percentage gradually increases up to around $98\%$. We further
quantitatively estimate the average number of Hessian entries each
individual Hessian entries (referred as \textbf{coupling set size})
are correlated with by computing the averages (over all Hessian
entries) of the number of non-zero covariances, skewnesses, and excess
kurtoses respectively throughout training, and the result is given in
\cref{fig:control_parameter_curve}. As shown by the light red line in
\cref{fig:control_parameter_curve} at the end of the training, for a
given Hessian entry, of overall $8.68 \cdot 10^{12}$ possible coefficients
between it and the rest of the entries---the network contains $N = 8.68
\cdot 10^{12}$ number of parameters---it only correlates $3.39 \cdot 10^{-4}$ of
them. Thus, each Hessian entry has a {\it very large absolute}
coupling set size and a {\it very small relative} coupling set size
compared with all the possible entries; the former encodes dependency
among data, and the latter shall enable adaptability.

\begin{figure*}[t]
  \centering
  \subfloat[\label{fig:eigenspectrum}]{%
  \includegraphics[width=0.45\linewidth]{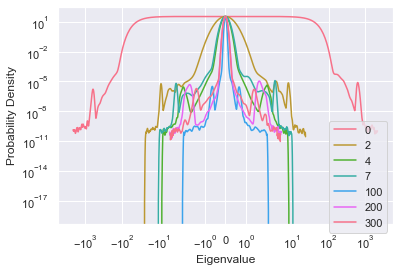}%
  }
  \subfloat[\label{fig:loss}]{%
  \includegraphics[width=0.49\linewidth]{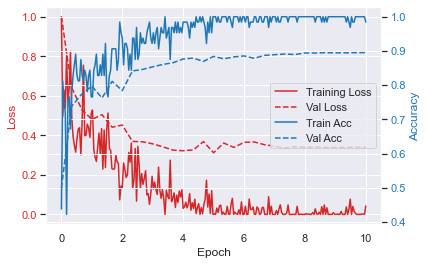}%
  }
  \caption{Plasticity phase, extended symmetry breaking and benign
pathways of DNNs. Hessian eigenspectrum is also a neuron assembly, and
under assumptions that characterize abundance of circuit symmetries,
is theoretically (through a theorem that analyzes the statistical
hierarchical many-body interaction among circuits) and experimentally
found to be self-similar as the gradients do as well, as shown in
fig. a. Such adaptive symmetry of eigenspectrum throughout training
implies that despite the circuit symmetries that break to decrease the
risk, the coarse-grained behaviors of the circuit symmetries are
macroscopically (system-widely) stable, and according to the
epistemology of physics, is a phase of the system, and is referred as
the plasticity phase. The plasticity phase of DNNs is an extended
adaptive-symmetries breaking process extended over time where circuit
symmetries continually break to reduce informational perturbations and
encode information from the environment. In the process, all
stationary points are saddle points, and benign pathways exist such
that by following gradients, zero risk could be reached, as shown in
fig. b.  To clarify, a subtlety exists in the problem setting
(introduced in \cref{sec:problem-setting-intro}) of this work, and
strictly speaking it is not that the eigenspectrum of Hessian is
symmetric but a matrix in a decomposed form of Hessian. {\bf (a)}
Adaptive symmetry of Hessian eigenspectrum throughout DNN
training. The numerically computed eigenspectrum of Hessian of the
12-layer VGGNet (cross entropy loss replaced by hinge loss) being
trained on modified CIFAR10 datasets for binary classification. The
numbers in the legends are epoch numbers in training. Note that the
axes are in logarithmic scale.  The eigenspectrum is computed with the
Lanczos spectrum approximation algorithm
\cite{Lin2013,Papyan2018,Ghorbani2019} (details in
\supps H). {\bf (b)} The loss (empirical risk)
and performance (accuracy) curves throughout DNN training. The loss
measures the gradually decrease of the informational perturbations
induced by examples in the training set, and the accuracy measures the
emergence of functional specialization that does classification on the
CIFAR dataset. Their change results from the breaking of circuit
symmetries.}
 \label{fig:exp}
\end{figure*}

\subsubsection{Plasticity phase, extended symmetry breaking, and benigh pathways of DNNs}
\label{sec:plast-phase-benigh}

We formulate those experimentally observed coarse-grained behaviors in
hierarchically large DNNs as assumptions, which could be appreciated
as conditions on coarse-grained behaviors in statistical physics
referred as control parameters such as temperature, but are more
heterogeneous because of the heterogeneity of circuits. And under
these assumptions, further analysis of Hessian reveals a phase that is
referred as the {\it plasticity phase}---thus, the parameters in the
assumptions would be appreciated as preconditions on control
parameters that identify a region of the parameter space that is a
phase, and interested readers could find the discussion in the
supplementary. Meanwhile, the phase is both a phase of a DNN, where
the DNN could continually decreases risk, and an {\it extended
symmetry-breaking} process, where the symmetries are continually being
broken during the self-organizing process. As a result, {\it benign
pathways} exist that lead to zero risk. We discuss these results next.

To begin with, we briefly contextualized the technique used in the
analysis.  The assumptions characterize a phenomenon of disorder at
the assembly level, which is mathematically equivalent to the disorder
at the field level in statistical physics. And it is not surprising
that we could reuse the techniques developed there to analyze the
disorder here. More specifically, sparsely correlated random matrices
are being studied actively in the random matrix community, and in our
setting---which is explained in \cref{sec:problem-setting}, and could be
roughly understood as the setting that DNNs do binary classification
with the hinge loss---the Hessian belongs to a class of random matrices
known as the Wigner-type matrix. The interests of Wigner-type matrices
began with its ability in modeling the nuclei of heavy atoms, which is
a disordered physical system that consists of a large number of
interacting subatomic units (i.e. nucleons)
\cite{Wigner1957,Guhr1998}. Though interaction exists among the
subatomic units, the mathematical difficulty in characterizing random
matrices with interaction/correlations is significant, and the early
works in field, e.g., the Wigner matrix \cite{Wigner1957}, ignored the
correlation among units. We shall build on recent advances
\cite{Helton2007,Erdos2017,Alt2018a} that allow the correlation among
matrix entries.

Under the preceding assumptions, we present a theorem that
characterizes a concentration-of-measure phenomenon, and states that
for a {\it hierarchically large} DNN, the eigenspectrum of its Hessian
is symmetric w.r.t. $y$-axis throughout training; that is, the
coarse-graining given in \cref{eq:dHd_circuit-pre} is self-similar to
the circuit symmetry of basis circuits. More specifically, the size of
a DNN is characterized by the {\it number of parameters}, which
depends on both {\it width} and {\it depth}. For example, let $n_l$
denote the width of a DNN (of depth $L$) at layer $l$, then, it has $N
= \sum_{l=1}^{L}n_{l-1}n_{l}$, or $N = Ln_0^2$ (assuming $\forall l \in \bb{L},
n_{l-1} = n_{l} = n_0$), number of parameters, which is a function of
both depth and width. This hierarchical largeness has taken the mantle
of {\it overparameterization} in the literature, and their
relationship is further discussed with related works in
\supps A D. Informally, we have
\begin{displaymath}
  \forall \lambda \in \bb{R}, \mu_{\bv{H}}(\lambda) \approx \mu_{\bv{H}}(-\lambda),
\end{displaymath}
where $\mu_{\bv{H}}$ denotes the eigenspectrum (i.e., probability
density distribution of eigenvalues) of $\bv{H}$---the formal version is
a probability bound in a theorem given in \supps B H 5.  The symmetry
of eigenspectrum implies that despite the broken circuit symmetry in
the self-organizing process, the order (i.e. the ability to decrease
risk) is stable because stationary point where the gradients (order
parameter) is zero could only be saddle points, and the risk decrease
would not be trapped in local minima. Therefore, in the regimen where
circuit symmetry dominates over broken circuit symmetry, the risk
could be minimized until its lower bound is reached, and for a well
designed loss function, its the lower bound is zero. In other words,
zero risk could be reached by following gradients through the
feedback-control loop of a DNN. Furthermore, a lower risk concentrates
eigenvalues more towards zero.

Experimentally, the stability of symmetries can be seen from
\cref{fig:eigenspectrum}, which shows that the eigenspectrum of the
DNN's Hessian is symmetrically distributed against $y$-axis throughout
training. \Cref{fig:eigenspectrum} also shows that the convergence of
the eigenspectrum to zero as training progress. Meanwhile, the broken
symmetries also macroscopically manifests as the decrease of risk as
training progress. This can be seen in \cref{fig:loss}, where the
empirical risk (i.e., loss) converges to zero and the training
accuracy converges to $1$.

\paragraph{Plasticity phase.}
Therefore, the conditions in the assumptions demarcate a region of the
state/weight/parameter space, where an abundant reservoir of circuits
with circuit symmetry exist (along with broken circuit symmetries that
encode information of the dataset), and thus the order parameter is
stably nonzero when information perturbations (novel examples) exist.
And as discussed in \cref{sec:adapt-symm-feedb}, unlike physical
systems, where the invariant are invariants of invariants, the
invariant of DNNs is the invariant of variants. Thus, the region of
state space where such stability of the coarse-grained effect of
circuit symmetries exists as a phase of DNNs.  We refer the phase as
the {\bf plasticity phase}.

\paragraph{Extended symmetry breaking.}  Meanwhile, the phase is {\it
both} a phase of a DNN, where the DNN could continually decreases
risk, and an {\bf extended symmetry-breaking} process, where the
symmetry is continually being broken during the self-organizing
process. Although the adaptive symmetries of assemblies hold
throughout the training, a subset of the circuits in the network
transit from a symmetric distribution to outputs that correspond to
zero risk of examples. That is, the excited perturbations gradually
converge to zero, because the informational patterns are encoded in
the network by circuits with broken circuit symmetries. In the Hessian
spectrum, this manifests as its gradual convergence to a zero function
because the examples in the training set are mapped to zero risk,
which has a zero Hessian matrix. And at the end of the training, zero
risk is reached.

\paragraph{Benign pathways.}  Consequently, in the plasticity phase,
by following the gradient---that is, by continually absorbing to
novel-example excitation by breaking the sufficient reservoir of
circuit symmetries through the feedback-control loop introduced in
\cref{sec:adapt-symm-feedb}---the system could decrease the risk without
being trapped in local minima, and eventually reach zero risk given a
(fixed) dataset. We refer the pathways found by gradients in this
phase as {\bf benign pathways}. And such pathways suggest an
explanation on the optimization power of DNNs.

To clarify lastly, the plasticity phase does not automatically hold
for any large DNNs processing any datasets. It is an intricate
interaction between the neurons and the data, and the generalization
to other network size, architectures or datasets should be studied
case by case. It is desirable to derive efficient algorithms to ensure
the assumptions are held online during DNN training, but this is
considered future works.

\subsection{Theoretical setting}
\label{sec:problem-setting}

Lastly, we give the detailed theoretical settings of this work.

\begin{enumerate}[leftmargin=0cm]
\item We investigate multilayer/deep neural networks that are
hierarchically large. As noted previously, we study complex systems,
and for complex systems irregularities of individual units are only
subdued into the ordered behaviors of the aggregated whole when the
number of units are large enough. To let the concentration-of-measure
phenomena discussed manifest, the networks need to be hierarchically
large. This hierarchically largeness is roughly known as the {\it
overparameterized regimen} in the literature, however, critical
difference exists---as we use the term ``hierarchical largeness''
instead of overparameterization---and is discussed in
\supps A D, where we discuss related works.
\item The study is specifically for DNNs with ReLU activation
function, a feedforward architecture, and without biases in each
layer's parameters, which also include convolutional DNNs (CNNs)
\cite{Lecun1998}---though the formalism is stated for multilayer
preceptron, CNNs mathematically are MLPs with sparse connection and
shared weights. As introduced in \cref{sec:bayes-defin-dnns}, ReLU is
a form of degenerated Monte Carlo sampling. The formalism could
generalize to some activation functions that are variants of ReLU,
e.g., Swish \cite{Ramachandran2017} and skip-connection architecture
like ResNet \cite{He2016}, but they are not pursued in this work and
are considered future works. Meanwhile, this formalism does not apply
for activation function like Sigmoid or Tanh: the circuit symmetries
do not hold for Sigmoid or Tanh. This fact also corroborates with the
fact that DNNs with ReLU are much easier to train than those
activation functions. Some further discussion on activation function
and circuit symmetries could be found in \supps A D 2,
where we discuss empirical works that study the risk landscape of
DNNs. In the experiments, no biases are used.
\item We study DNNs that do binary classification.  As introduced
previously, DNNs are studied by analyzing the feedback-control loop
between the hierarchical circuits and coarse-grained
variables. Theoretically, the coarse-grained variable could be any
form, however, to investigate the principle underlying DNNs without
unnecessary complications, we study DNNs that do binary
classification. In this case, the coarse-grained variable is a single
output neuron at the top layer that classifies macroscopic patterns
composed by a large number of microscopic units, e.g., pixels. And the
loss function composed with DNNs belongs to a class of loss functions
whose representative is the hinge loss.  The class of loss functions
are formally characterized in \supps B C 8. And the class of
functions do not include quadratic loss, or cross entropy loss. Such
restriction was firstly studied in \citet{Choromanska2015}
(cf. \supps A D 4 for related-works discussion from the
statistical-physics approach). This is a beachhead problem that is
simple yet nontrivial: this setting is the first problem of two
interdependent problems that make up the general setting.  For
interested readers, the problem decomposition is given in
\supps G A 3, and furthermore, how the results in this
work could potentially generalize to all kinds of loss functions is
also given in \supps G A 2.
\item We do not made any assumptions on input data except that they
are normalized to the order of magnitude $\Theta(1)$; or, putting it in
another way, the theoretical results apply to practical data. To
elaborate, the assumptions introduced in
\cref{sec:coarse-grain-effect-3} characterize a certain interaction
between neurons and data such that when processing the data, certain
adaptive symmetries and diversity exist in the neuron assemblies. As
discussed in \cref{sec:disc-compl-from}, the assumptions could be
understood as certain complexity-matching between the dataset and the
network.
\end{enumerate} Overall, except for relatively the simple loss
function, the setting is the setting that is widely used in practice.

\section{Discussion: complexity from adaptive-symmetries breaking}
\label{sec:disc-compl-from}

The results introduced from \cref{sec:umwelt} to
\ref{sec:plast-phase-benign} compose an iteration of scientific
inquiry where falsifiable hypotheses are developed from a formalism of
the phenomenon and then are experimentally validated. Yet, it leaves
many problems not investigated. In \supps C, we shall
venture into the creative side of scientific inquiry, and try to make
sense of the this work in relation to science in general. We briefly
remark the contents there in this section for interested readers.

In \supps C, we clarify understudied concepts in the
extended symmetry-breaking process of DNNs, and point out the
connection between adaptive-symmetries breaking and complexity. More
specifically, the plasticity phase characterized in this work looks
like a paradox: in the plasticity phase, a DNN both possesses the
stability of coarse-grained circuit symmetries, and the continual
breaking of microscopic circuit symmetries.  The paradoxical behaviors
of DNNs could be appreciated as the paradigmatic-shift behaviors of
organized complex systems in the sense of Kuhn's \cite{Kuhn:1970}
paradigm shift in science.  Particularly, DNNs belong to the class of
systems that are known as {\it organized complexity}
\cite{Scientific2016}, and are referred as {\it complex systems}
\cite{Kornberger2003,Amderson1972,10.5555/1614219,erdi2007complexity,Nicolis:2012:FCS:2331101,thurner2018introduction}.
And this superficially paradoxical behavior emerges from the increased
potential complexity of the DNN system. Furthermore, the increase of
potential complexity also requires qualitative generalization of
certain foundational concepts in statistical physics, such that they
could be applied to the DNN organized complex system. In
\supps C, we expand on the discussion here. We first put
DNNs in a spectrum of phenomenological models that characterize
increasingly complex systems in nature, and it is from this complexity
increase, that the paradoxical behaviors emerge. Then, we discuss the
generalized (from physics) concepts of criticality and phase
transition of DNNs. Lastly, we discuss the self-organization of DNNs
from the perspective of complexity increase. We given the outline in
the following.

\paragraph{Phenomenological models with increasing complexity: spin
glasses and DNNs.}  First, in \supps C A 1, we compare
the training process of DNNs with the frustration process of spin
glasses, and in \supps C A 2, point out that the
paradox of breaking and stable symmetries could be appreciated by
putting spin glasses and DNNs in a spectrum of models with increasing
potential complexity: it is the vastly increased potential
complexity—which we mean by a hierarchically large DNNs with abundance
of reservoir of circuits with circuit symmetries—enables a diachronic
``frustration'' process where circuit symmetries continually break,
and makes the self-organizing of DNNs both a symmetries-breaking
process and a phase with stable symmetries.

\paragraph{Extended criticality.}  Second, the study of symmetry
breaking in this work might be confusing without mentioning the
criticality, which is a concept associated with symmetry breaking in
physics. In \supps C B, we clarify the concept of
criticality in DNNs. In physics, symmetry breaking is
associated with power-law criticality. However, as the self-organizing
process extends from a frustration process to a diachronic learning
process, the criticality of DNNs is also spatially and temporally
extended.  More specifically, the concept of criticality needs to be
reconceptualized as {\it extended criticality} \cite{Longo2013} in
theoretical biology, which is introduced in
\supps C B 1 and states that the criticality of
biotic systems lies in an extended region of the state space instead
of singular points (as in the physical systems). In
\supps C B 2 the self-organizing of DNNs is
characterized as a process where the system computes at extended
criticality to encode information and decrease uncertainty. The
concept of extended criticality was developed further from the concept
of {\it edge of chaos}
\cite{packard1988adaptation,Kauffman1991,Gramb2005}, which under the
context of biology refers to the phenomenon that the biotic systems
are in a diachronic evolving process in regions of the state space
that is between chaos and equilibrium. The concept that a DNN computes
at edge of chaos is not novel, and existing works
\cite{Poole2016,Feng2019,Zhang2020a,Zhang2021} have interpreted the
training of DNNs as such. Here, although the relationship between
circuit-symmetry breaking and edge of chaos has not been discussed
before, the discussion is mostly to clarify the criticality in the
symmetry-breaking of DNNs. In \supps C B 3, we show
that, mostly empirically, the a DNN self-organizes in the extended
critical regimen during training.

\paragraph{Extended phase transition.}  Third, as an extended
symmetry-breaking process, theoretically, the reservoir of circuit
symmetries could eventually be exhausted, therefore, in
\supps C C, we shall discuss the self-organizing
process from the perspective of extended phase transition. We have
studied the region of the phase space where circuit symmetries coexist
stably with broken circuit symmetries. As the symmetry-breaking is
kept being broken---when more complex datasets are being processed---a DNN
theoretically could run out of circuit symmetries. Therefore, we
tentatively classify the self-organizing process of DNNs into two
phases: in addition to the plasticity phase with stable circuit
symmetries, the other phase contains a paucity of adaptive symmetries
that could easily be unable to decrease the errors of novel examples,
and is referred as the {\it frustration phase}—note that the
frustration here does not mean the system is trapped in local minima
that are not global minima, but simply be of nonzero risk.  Two phases
are quantitatively characterized by the plasticity order parameter and
Hessian: in the plasticity phase, the order parameter is nonzero (when
novel examples still exist) and the Hessian's spectrum is symmetric,
and in the frustration phase, the order parameter is zero and the
Hessian is likely positive semidefinite. Compared with the phase space
of statistical physics, where the trajectories can be visualized as a
function of order parameter against scalar control parameters thanks
to the homogeneous symmetries, the phase space of DNNs is a
high-dimensional manifold embedded in the ambient state/weight
space. We shall discuss a toy case of the frustration phase in
\supps C C 1 and remark the phase transition between
the two phases in \supps C C 2, mostly to clarify that
the extended symmetry-breaking should also be understood as an
extended phase transition.

\paragraph{Complexity from adaptive-symmetries breaking.}  Lastly, in
\supps C D, we discuss that symmetry breaking and
complexity is like a duality: the accumulation of a large number of
broken adaptive symmetries manifest as the complex behaviors of a
system. Therefore, we point out a shift of perspective where the
self-organizing of DNNs could be studied from the perspective of
breaking symmetries to that of increasing complexity. More
specifically, in this work, we give a characterization of the risk
minimization of DNNs as a process where the system self-organizes to
minimize uncertainty by breaking a reservoir of adaptive-symmetries in
the feedback-control loop between it and the environment where it is
embodied. In doing so, we identify a plasticity phase of DNNs that
suggests an explanation of the optimization power of DNNs. However,
symmetry-breaking emphasizes the physical perspective, and DNNs'
self-organizing is also a computing process, where information about
the environment is gradually encoded into the circuits. And this
computing perspective could be qualitatively described as increasing
complexity, referring the increasing complex patterns recognizable by
DNNs. Therefore, we summarize the process as complexity from adaptive
symmetries as follows.
\begin{quotation}
  DNNs are able to increase the potential complexity of the system
plastically, and when the potential complexity of a system is larger
than the complexity of a dataset, a DNN could absorb informational
perturbations from the environment and self-organize into a functional
structure that reaches a goal with zero training errors measured by a
ncertain surrogate risk.
\end{quotation}

\end{document}